\documentclass[10pt,final,journal]{IEEEtran}

\usepackage{amsmath,amssymb,amsfonts}
\usepackage{graphicx}
\usepackage{textcomp}
\usepackage{algorithm}
\usepackage[noend]{algpseudocode}
\usepackage{setspace}
\usepackage{enumitem}

\DeclareMathOperator{\argmax}{arg\,max}
\usepackage{tabularx}
\usepackage{multirow}
\usepackage{multicol}
\usepackage{diagbox}
\usepackage{adjustbox}

\usepackage{caption}
\usepackage{subcaption}
\usepackage{comment}
\usepackage{placeins}

\usepackage{xcolor}

\usepackage[numbers,sort&compress]{natbib}

\usepackage{array}
\newcolumntype{P}[1]{>{\centering\arraybackslash}p{#1}}

\begin{document}

\title{Attention-based Adversarial Robust Distillation in Radio Signal Classifications for Low-Power IoT Devices}

\author{Lu Zhang$^{1}$, Sangarapillai Lambotharan$^{1}$, Gan Zheng$^{2}$, Guisheng Liao$^{3}$, Basil AsSadhan$^{4}$, Fabio Roli$^{5}$ \\$^{1}$Wolfson School of Mechanical, Electrical and Manufacturing Engineering, Loughborough University, Loughborough, UK\\$^{2}$ School of Engineering, University of Warwick, Coventry, CV4 7AL, UK
\\$^{3}$School of Electronic Engineering, Xidian University,
Xi’an 710071, People’s Republic of China\\$^{4}$Department of Electrical Engineering, King Saud University, Riyadh, Saudi Arabia\\$^{5}$Department of Informatics, Bioengineering,
Robotics, and Systems Engineering, University of Genova, 16145 Genoa, Italy}

\IEEEtitleabstractindextext{
\begin{abstract}
Due to great success of transformers in many applications such as natural language processing and computer vision, transformers have been successfully applied in automatic modulation classification. We have shown that transformer-based radio signal classification is vulnerable to imperceptible and carefully crafted attacks called adversarial examples. Therefore, we propose a defense system against adversarial examples in transformer-based modulation classifications. Considering the need for computationally efficient architecture particularly for Internet of Things (IoT)-based applications or operation of devices in environment where power supply is limited, we propose a compact transformer for modulation classification. The advantages of robust training such as adversarial training in transformers may not be attainable in compact transformers. By demonstrating this, we propose a novel compact transformer that can enhance robustness in the presence of adversarial attacks. The new method is aimed at transferring the adversarial attention map from the robustly trained large transformer to a compact transformer. The proposed method outperforms the state-of-the-art techniques for the considered white-box scenarios including fast gradient method and projected gradient descent attacks. We have provided reasoning of the underlying working mechanisms and investigated the transferability of the adversarial examples between different architectures. The proposed method has the potential to protect the transformer from the transferability of adversarial examples.

\end{abstract}

}

\maketitle

\IEEEdisplaynontitleabstractindextext
\IEEEpeerreviewmaketitle

\section{Introduction}
The Internet of Things (IoT) and mobile networks are evolving rapidly to fulfil the need for ultra reliable and low latency performance, seamless connectivity, mobility, and intelligence \cite{cui2021integrating, mu2021machine, zhang2020device, wang2019energy}. It is estimated that over 50 billion devices are wirelessly connected to the internet, which can sense their surroundings and offer high-quality services. The explosive growth of IoT devices demands efficient management of already scarce radio spectrum, which is very challenging particularly in a non-cooperative communication environment. As a result, classifying modulation types at the receiver under non-cooperative communication conditions becomes a critical task. Automatic modulation classification (AMC) is proposed which plays a key role in wireless spectrum monitoring by performing modulation classifications possibly without prior knowledge of the received signals or channel parameters \cite{weber2015automatic, clancy2007applications}. It also plays an important role in wireless spectrum anomaly detection, transmitter identification and radio environment awareness, consequently improving radio spectrum usage and the context aware intelligent decision making in autonomous wireless spectrum monitoring.  

Traditionally, AMC was mainly achieved by carefully extracted features (such as higher order cyclic moments) by experts and certain classification criteria \cite{wu2008novel, ramkumar2009automatic, park2008automatic, swami2000hierarchical,soliman1992signal}. These existing feature-based methods are easy to implement in practice, however, hand-crafted features and hard-coding criteria for AMC make scaling to new modulation types challenging. Recently, due to the superior performance of deep learning, many researchers have resorted to various deep neural network (DNN) architectures for AMC \cite{o2016convolutional, o2018over, krzyston2020high, liao2021sequential, ramjee2019fast, hou2022multi, dong2022lightweight, fu2022automatic}. For example, a convolutional neural network (CNN) was used for AMC in \cite{o2016convolutional}. Later convolutional long short-term deep neural networks (CLDNN), long short-term memory neural networks (LSTM), and deep residual networks (ResNet) were proposed to improve the classification performance \cite{ramjee2019fast}. A complex CNN was proposed in \cite{hou2022multi} for the identification of signal spectrum information. A spatio-temporal hybrid deep neural network was proposed in \cite{dong2022lightweight} for AMC which is based on multi-channels and multi-function blocks. Furthermore, to reduce the communication overhead, the authors of \cite{fu2022automatic} proposed an innovative learning framework which is based on the combination of decentralized learning and ensemble learning. With the great success of transformer in the computer vision area \cite{dosovitskiy2020image, wang2021pyramid, graham2021levit}, the work in \cite{hamidi2021mcformer} has successfully applied transformers in AMC which shows considerable performance improvement compared to the state-of-the-art techniques. 

Despite its superior performance of DNN, several recent research works have pointed out that DNNs are vulnerable to adversarial examples, which are imperceptible and deliberately crafted modifications to the input that result in misclassifications \cite{goodfellow2014explaining}. Adversarial examples have been proven to be effective in terms of hindering operation of several machine learning applications, such as face recognition \cite{sharif2016accessorize}, object detection \cite{xie2017adversarial}, semantic segmentation \cite{hendrik2017universal}, natural language processing \cite{jia2017adversarial}, and malware detection \cite{hu2017generating}. Notably, adversarial examples using fast gradient methods (FGM) have been shown to reduce the classification accuracy in AMC \cite{sadeghi2018adversarial, zhang2021countermeasures}. Recently, we have shown a transformer-based AMC is also vulnerable to adversarial examples using a projected gradient descent (PGD) method \cite{lu2022Icassp}.

\begin{figure}[ht]
\centering
\includegraphics[width=\columnwidth]{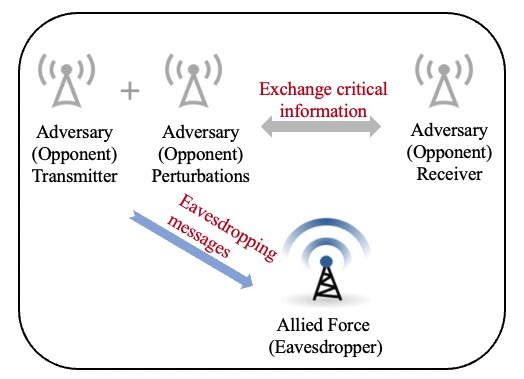}
  \caption{A military scenario of the adversarial examples in modulation classification.}
  \label{fig:eavesdropper}
\end{figure}

In practice, AMC can be applied to both military and civilian scenarios. In a networked battlefield, important information may be shared using radio signals by the units of each adversary (opponent transmitter and receiver as indicated in Figure \ref{fig:eavesdropper}). The allied forces (playing the role of eavesdropper in this scenario) can employ AMC to determine the modulation used to eavesdrop information transmitted between the adversary units (opponents). To deter the allied forces from eavesdropping messages, small perturbations (adversarial perturbations) can be applied to the communication signals by the opponents such that the modulation discovery executed by the allied forces eventually fails. The modulation can still be discovered by the allied forces, however in this case, an AMC system which is robust against adversarial attacks should be applied as proposed in this paper. To the best of our knowledge, this work is the first to propose a defense for the transformer-based AMC in the literature. The proposed defense will be used to protect the allied forces from the adversarial perturbations such that the allied forces could successfully discover the modulation. While the transformers offer superior performance, it comes with a price in terms of large model size and computational complexity, which may limit reaping its benefits in applications that rely on low-power sensors and IoT networks \cite{zhou2019access, liu2018deep, zhou2015energy, tang2019future} with possibly a low memory size. Therefore we propose a novel compact transformer-based defense for low-power IoT devices based on distillation of knowledge through the adversarial attention map, which is a critical element for the transformer and the details are given in Section \uppercase\expandafter{\romannumeral3}.

\setlength{\textfloatsep}{0.6mm}

There are several countermeasures to defend DNN against adversarial examples \cite{papernot2016distillation, jia2019comdefend, kurakin2016adversarial, ma2018characterizing, madry2017towards}. Among these defenses, adversarial training (AT) has been shown to be the most powerful method \cite{athalye2018obfuscated, croce2020reliable}. It is a data augmentation technique which augments adversarial examples during the training of DNN.  While AT offers superior performance in large transformers, such benefits are not apparent in compact transformers as shown in Table \ref{tab:different modulation scheme}, which shows  robustness (classification accuracy in the presence of attacks), especially when the PNR (i.e., the ratio of the adversarial perturbation power to the noise power) is high. Therefore, we focus on transferring the robustness from a large transformer to a compact transformer. For DNN, there have been some works in computer vision applications that considered transferring robustness from a teacher model to a student model. For example, an adversarially robust distillation (ARD) method for ordinary DNN was proposed in \cite{goldblum2020adversarially} to distill robustness into small student networks during knowledge distillation. A reliable introspective adversarial distillation (IAD) technique was proposed in \cite{zhu2021reliable} where students partially trust their teachers instead of trusting them fully. The work in \cite{chen2020robust} used both robust-trained and standard-trained self-teachers, which we call adversarial knowledge distillation (AKD). Furthermore, a robust soft label adversarial distillation (RSLAD) method was proposed in \cite{zi2021revisiting} which fully exploits the robust soft labels obtained by a robust teacher model to teach the student’s learning. However, these methods consider only the logits (i.e., representations of the penultimate layer of the DNN) information from the robust teacher models which may not be sufficient to maintain robustness. In this paper, based on a unique architecture of the transformer-based neural network, we propose an ATtention-based Adversarial Robustness Distillation (ATARD) method for low-power IoT devices to transfer the robustness onto a compact transformer model by learning the adversarial attention map from a robust large transformer model. The proposed ATARD has a better smoothness (i.e., less sensitive to perturbations) than AT, ARD, IAD, AKD, and RSLAD as demonstrated in Section \uppercase\expandafter{\romannumeral3}. Our key contributions are:
\begin{itemize}
  \item For the first time in the literature we propose a defence against adversarial attacks for transformer-based AMC.
  \item We propose an attention-based adversarial robustness distillation method to transfer the robustness onto a compact transformer model by learning the adversarial attention map from a robust large transformer model. To the best of our knowledge, this is the first work to distill robustness by transferring the adversarial attention map onto small networks.
  \item Considering white-box attacks, we show the performance advantage of our proposed method compared to the state of the art techniques including AT, ARD, IAD, AKD, and RSLAD. Furthermore, the transferability of the adversarial examples among different architectures is shown and the robustness of ATARD is established.
\end{itemize} 

The rest of the sections are arranged as follows: Section \uppercase\expandafter{\romannumeral2} reviews the related works. The proposed
methodology is presented in Section \uppercase\expandafter{\romannumeral3} followed by the results and discussions in Section \uppercase\expandafter{\romannumeral4}. Finally conclusions are drawn in Section \uppercase\expandafter{\romannumeral5}.

\begin{table}[]
\caption{Accuracy of the AT trained compact transformer as compared to the AT trained large transformer against PGD attacks for a wide range of PNR values.}
\label{tab:different modulation scheme}
\begin{center}
\scalebox{0.9}{
\begin{tabular}{|c|l|l|l|}

\hline
PNR                            & -30~dB  & -20~dB  & -10~dB   \\ \hline
AT trained large transformer & 86.3\% & 82.2\% & 63.1\%  \\ \hline
AT trained compact transformer     & 78.8\% & 74.54\% & 50.79\% \\ \hline
\end{tabular}}
\end{center}
\end{table}

\section{Related Work}
First we elaborate on the related techniques including AT, ARD, IAD, AKD, and RSLAD. AT can be traced back to Goodfellow et al. \cite{goodfellow2014explaining}, in which the clean images and the corresponding adversarial examples were mixed into every mini-batch for training. The loss function of AT is expressed as:
\begin{equation}
\label{equ:loss function of AT}
\alpha CE(f_{\theta }(\mathbf{x}),y)+(1-\alpha) CE(f_{\theta }(\mathbf{x}^{adv}),y),
\end{equation}
where $CE(\cdot )$ is the cross entropy loss. $\theta$ is the parameter of the network $f(\cdot)$, $\mathbf{x}$ means the clean (benign) samples without adversarial perturbation, $\mathbf{x}^{adv}$ is the adversarial counterparts of the corresponding $\mathbf{x}$, and $y$ is the true label of the sample. $\alpha$ accounts for the relative importance between the clean sample loss and the adversarial sample loss. We choose $\alpha=0.5$ as suggested in \cite{goodfellow2014explaining}. The first term in \eqref{equ:loss function of AT} is to guarantee the classification accuracy for benign samples, and the second term is to maintain the accuracy when the network is exposed to adversarial samples. The optimization function of the procedure for generating the adversarial samples $\mathbf{x}^{adv}$ is the following:
\begin{equation}
\label{equ:loss function1 of AT1}
\mathbf{x}^{adv}=\underset{\left \| \mathbf{x'}-\mathbf{x} \right \|_{p}\leq \varepsilon }{\operatorname{argmax}} CE(f_{\theta }(\mathbf{x}),y)
\end{equation}
We use a 3 step-PGD attack as suggested by \cite{madry2017towards}. The PGD attack is the strongest attack utilizing the local first order information about the network. The value of $\alpha$ and the generation of $\mathbf{x}^{adv}$ remain the same for the rest of the techniques unless stated otherwise. 

ARD was proposed to distill robustness from a robust teacher to a small student network. ARD is analogous to AT but in a distillation setting. During AT, a DNN is used to predict the true label when the input to the DNN is exposed to adversarial perturbation. While in ARD, given a robust (AT-trained) teacher model $T$, a student model $S$ would aim to match the teacher network's output when exposed to an adversary. At the same time, the loss function between the output of the student model and the ground truth label was considered to balance the natural accuracy. The loss function of the ARD is as follows \cite{goldblum2020adversarially}:
\begin{equation}
\label{equ:loss function of ARD}
\alpha t^{2}KL(S^{t}_{\theta }(\mathbf{x}^{adv}), T^{t}(\mathbf{x}))+(1-\alpha )CE(S^{t}_{\theta }(\mathbf{x}),y),
\end{equation}
where $KL(\cdot)$ is KL-divergence loss and $t$ means temperature. The logits of both the teacher and the student networks are divided by the temperature term $t$. In this work, we set $t=1$. 

In terms of adversarial robustness, the authors of \cite{zhu2021reliable} consider that the teacher models may become unreliable so that adversarial distillation may not work. This is because the teacher models are pre-trained on their own adversarial samples, and it is unrealistic to expect the teacher models to be reliable for each adversarial sample inquired by student models. Hence IAD was proposed in which the student trusts the teacher network only partially than fully. To be specific, given a query of an adversarial sample and its corresponding adversarial sample from the student network, IAD considers three different situations. First, if a teacher performs well at the adversarial sample, then its outputs (or soft labels) can be completely trusted. Second, if a teacher performs well at clean sample but not at adversarial samples, its output is partly trusted and the student network will rely on its own outputs. Otherwise, the student model will only trust its own outputs. The loss function of IAD is shown as below:
\begin{equation}
\label{equ:loss function of ARD}
\alpha KL(S^{t}_{\theta }(\mathbf{x}^{adv}), T^{t}(\mathbf{x}))+(1-\alpha )KL(S^{t}_{\theta }(\mathbf{x}^{adv}), S^{t}(\mathbf{x})),
\end{equation}
where $\alpha$ is a parameter which is introduced to balance the influence of the teacher network in IAD. As shown in \eqref{equ:loss function of IAD}, $\alpha$ is defined as the probability of the teacher network about the targeted label $y$ when queried by the adversarial data, and $\beta$ is used to sharpen the prediction probability.
\begin{equation}
\label{equ:loss function of IAD}
\alpha =(P_{T}(\tilde{\mathbf{x}}\mid y))^{\beta },
\end{equation}

In AKD, the authors propose to inject learned smoothing during AT in order to avoid the overfitting issue for the AT trained models. AKD utilizes both self-training and knowledge distillation to smooth the logits. Specifically, the teacher model is trained using two methods: standard training and robust training, which yields two teacher models. The loss function of AKD is expressed as follows:
\begin{equation}
\begin{aligned}
&~~~~~~~~~~~~~~~~~(1-\lambda _{1}-\lambda _{2})\cdot  CE(S_{\theta }(\mathbf{x}^{adv}), y)\\
&+\lambda _{1}\cdot KL(S_{\theta }(\mathbf{x}^{adv}), T_{at}(\mathbf{x}))+\lambda _{2}\cdot KL(S_{\theta }(\mathbf{x}^{adv}), T_{std}(\mathbf{x})),
\end{aligned}
\label{equ:loss function of AKD}
\end{equation}
where $T_{at}(\mathbf{x})$ is the AT-trained teacher network and $T_{std}(\mathbf{x})$ is the standard-trained teacher network. $\lambda _{1}$ and $\lambda _{2}$ are hyperparameters, and we use $\lambda _{1}=0.5$ and $\lambda _{2}=0.25$ following the default setting in AKD. 

Based on both AT and ARD, the authors of RSLAD \cite{zi2021revisiting} observed that the use of predictions of an adversarially trained model could improve the robustness. Hence a robust soft label adversarial distillation (RSLAD) method was proposed, which brings robust soft labels into its full play, i.e., the robust soft labels were used in both the loss function of RSLAD and the generation of the adversarial counterparts during training. The overall optimization is expressed as follows:
\begin{equation}
\label{equ:loss function of RSLAD}
\alpha KL(S^{t}_{\theta }(\mathbf{x}^{adv}), T(\mathbf{x}))+(1-\alpha )KL(S_{\theta }(\mathbf{x}), T(\mathbf{x})),
\end{equation}
where $T(\mathbf{x})$ are the robust soft labels obtained by the adversarially trained teacher model, and the adversarial samples $\mathbf{x}^{adv}$ are generated using:
\begin{equation}
\label{equ:loss function1 of RSLAD}
\mathbf{x}^{adv}=\underset{\left \| \mathbf{x}'-\mathbf{x} \right \|_{p}\leq \varepsilon }{\operatorname{argmax}} KL(S(\mathbf{x}),T(\mathbf{x})).
\end{equation}
No hard labels were used in the objective function of RSLAD and the most commonly used CE loss was replaced by KL-divergence loss to express the degree of distributional difference between the output probabilities of the two models.

\begin{figure*}[ht]
\centering
\includegraphics[width=\textwidth]{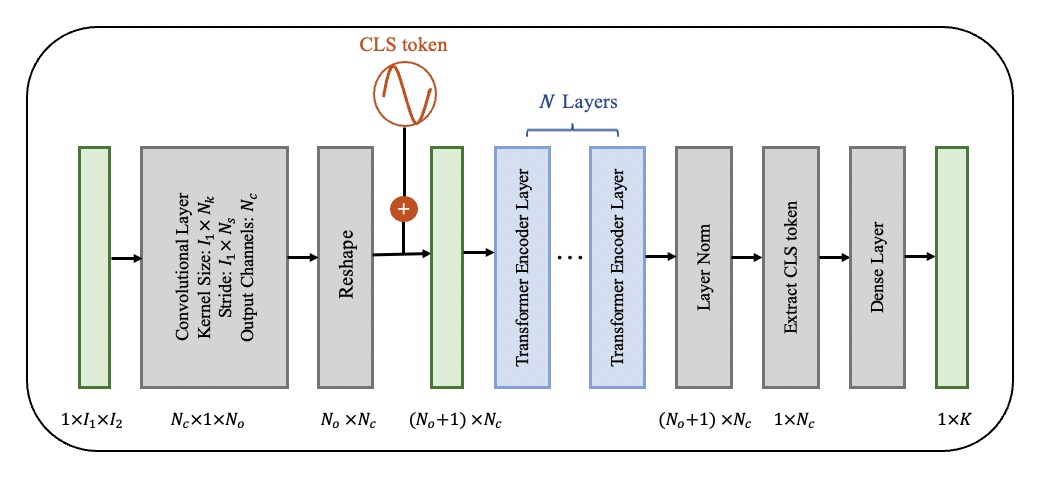}
  \caption{The architecture of the transformer-based teacher network \cite{lu2022Icassp}.}
  \label{fig:trannsformer}
\end{figure*}

\section{The proposed attention-based adversarial robustness distillation}
In this section, we provide details of our proposed attention-based adversarial robustness distillation (ATARD) method for low-power IoT devices. Two main differences between the proposed ATARD and the existing techniques AT, ARD, IAD, AKD, and RSLAD are as follows. The difference between transformers and CNNs is that, the self-attention in transformers not only extracts intra-patch features but also considers the inter-patch relations, Therefore, ATARD is based on the transformer networks, i.e., the teacher and student networks are both based on the transformer architecture. Second, instead of transferring the robustness through the prediction probability, the ATARD transfers the attention map obtaining inherent information from the teacher model to the student model. 

\subsection{The proposed ATARD}
Before delving into details, we first provide an introduction to the large teacher transformer model. We adopt the same transformer architecture as that used in \cite{lu2022Icassp} for the teacher network as shown in Figure \ref{fig:trannsformer}. The input to teacher transformer is such that I and Q components form a two-dimensional image of depth one ($1\times I_{1}\times I_{2}$ dimension). After going through a convolutional layer and a reshaping layer, this input signal becomes $N_{0}$ patch embeddings, where $N_{o}=\frac{I_{2}-N_{k}}{N_{s}}+1$, and each patch embedding has $1\times N_{c}$ (in our case $N_{c}=128$) dimension. Then a learnable embedding (CLS token) is prepended to the sequence of embedded patches, whose state at the output of the transformer encoder serves as the signal representation. The CLS token has the same dimension as each embedded patch and its parameters are learned during the backpropagation of the training process. Hence in total we have $(N_{0}+1)$ patch embeddings denoted as $\mathbf{z_{0}}$, and these patch embeddings are fed into $N$ (in our case $N=4$) consecutive transformer encoder layers. These encoder layers will not change the dimension of the input, i.e., the output of these four transformer encoder layers has the same dimension as the input patches ($(N_{0}+1)\times N_{c}$). After a layer normalization (LN) \cite{ba2016layer} process, the CLS token information is extracted and processed through a dense layer. Finally, a $1\times K$ vector is produced which represents the prediction probabilities of $K$ different classes for the input signal. 

\begin{figure}[ht]
\centering
\includegraphics[width=\columnwidth]{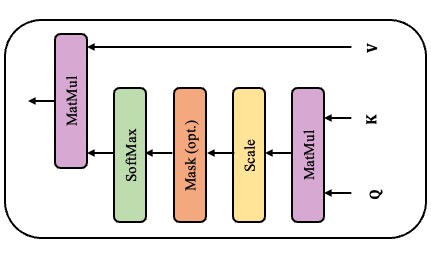}
  \caption{Scaled Dot-Product Attention \cite{vaswani2017attention}.}
  \label{fig:scaled attention}
\end{figure}

The encoder layer, as a key part of the transformer network, consists of two sub-layers. The first one is a multi-head self-attention (MSA) function, and the second is a position-wise fully connected feed-forward network (FFN). An LN is used around each of the two sub-layers, followed by a residual connection \cite{he2016deep}. Specifically, the output of the encoder layer $\mathbf{z_{n}}$ can be expressed as follows:
\begin{equation}
\label{equ:transformer encoder1}
\mathbf{z_{n}'}=\textup{SA}(\textup{LN}(\mathbf{z_{n-1}}))+\mathbf{z_{n-1}}, ~~~~~n=1,...N
\end{equation}
\begin{equation}
\label{equ:transformer encoder2}
~~~~\mathbf{z_{n}}=\textup{FFN}(\textup{LN}(\mathbf{z_{n-1}'}))+\mathbf{z_{n-1}'},~~~~~n=1,...N
\end{equation}
The standard self-attention (SA) mechanism, as shown in Figure \ref{fig:scaled attention}, is a function which converts a query and a set of key - value pairs into an output. As in \eqref{equ:attention1}, the output of SA is generated by calculating the weighted sum of the values $\mathbf{V}$ and the associated weight $\mathbf{A}$ assigned to each value. We denote the weight $\mathbf{A}$ as the attention map which is obtained as a function of the query $\mathbf{Q}$ and the corresponding key $\mathbf{K}$ with a dimension of $d_{k}$. Specifically, the weight $\mathbf{A}$ is calculated as the scaled dot products of the query with the keys, followed by using a softmax function as shown in \eqref{equ:attention2}:

\begin{equation}
    \label{equ:attention1}
     \textup{SA}(\mathbf{Q},\mathbf{K},\mathbf{V})= \mathbf{A}\mathbf{V}
\end{equation}

\begin{equation}
    \label{equ:attention2}
     \mathbf{A} = \textup{softmax}(\frac{\mathbf{Q}\mathbf{K}^{\textup{T}}}{\sqrt{d_{k}}})
\end{equation}
MSA is an extension of SA, in which SA operations are implemented $h$ times in parallel, and their outputs are concatenated and projected. Hence, the output of MSA can be expressed as:
\begin{equation}
    \label{equ:msa}
     \textup{MSA}(\mathbf{Q},\mathbf{K},\mathbf{V})= [\textup{SA}_{1}; \textup{SA}_{2};...;\textup{SA}_{h}]\mathbf{U}_{\textup{MSA}},
\end{equation}
where $\mathbf{U}_\textup{{MSA}}$ is a linear projection matrix.

\begin{figure}[ht]
\centering
\includegraphics[width=\columnwidth]{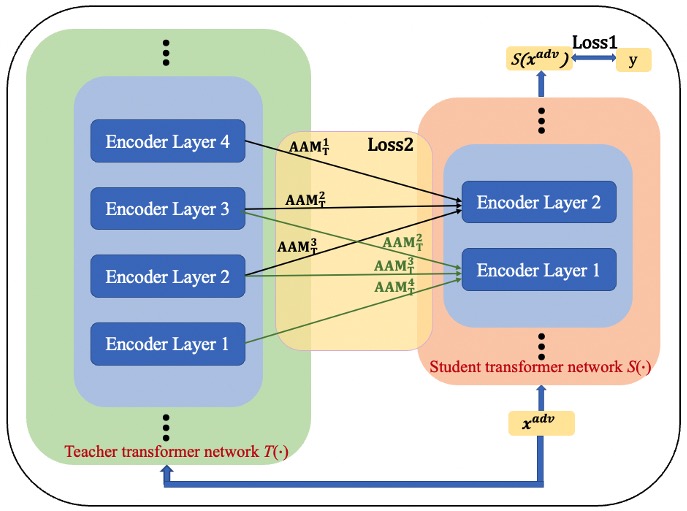}
  \caption{The architecture of the proposed ATARD method.}
  \label{fig:attention transfer}
\end{figure}

We now present the proposed ATARD method that transfers the robustness from a large teacher transformer network into a small student transformer model as shown in Figure \ref{fig:attention transfer}. The computational complexity is proportional to the total number of parameters of the network. To reduce the computational complexity, we use $N=2$ and $N_{c}=96$ for the small transformer network, and the total number of parameters has been reduced from $801,675$ to $230,699$. The calculation of the number of the parameters for both the teacher and the student transformer network is shown in the Appendix. The teacher transformer network is pre-trained using AT as in \eqref{equ:loss function of AT} and \eqref{equ:loss function1 of AT1}, then during the training of the student transformer networks, instead of only constraining loss function between the predicted output of the adversarial samples and true label $y$, we also force the adversarial attention map extracted from the teacher transformer network $T(\cdot)$ and the student transformer network $S(\cdot)$ as similar as possible. Specifically, during each training iteration of the student network, the adversarial samples $\mathbf{x}^{adv}$ are generated for each benign sample. Feeding $\mathbf{x}^{adv}$ into both teacher and student network, the adversarial attention map is extracted. For notational simplicity, we denote the adversarial attention map for four encoder layers of the teacher network as $\mathbf{AAM}_{T}^{1}$, $\mathbf{AAM}_{T}^{2}$, $\mathbf{AAM}_{T}^{3}$, $\mathbf{AAM}_{T}^{4}$, respectively. Similarly, the adversarial attention map for two encoder layers of the student is denoted as $\mathbf{AAM}_{S}^{1}$, $\mathbf{AAM}_{S}^{2}$. As mentioned above, an MSA is used in this work which means rather than implementing a single attention function, the query $\mathbf{Q}$ and the key $\mathbf{K}$, are linearly projected $h$ times with different, learned linear projections. Hence we have $h$ different queries and keys denoted as ${\mathbf{Q}_{1},...,\mathbf{Q}_{h}}$ and ${\mathbf{K}_{1},...,\mathbf{K}_{h}}$ respectively. Therefore, in terms of the $\mathbf{AAM}$ for both the teacher and the student network, we use the average of $h$ keys and queries. Hence, the $\mathbf{AAM}$ can be written as:
\begin{equation}
\label{equ:AAM}
\mathbf{AAM}=\sum_{i=1}^{h} \textup{softmax}(\frac{\mathbf{Q}_{i}\mathbf{K}_{i}^{T}}{\sqrt{d_{k}}})
\end{equation}

Then we force the $\mathbf{AAM}_{S}^{1}$ to learn from $\mathbf{AAM}_{T}^{1}$, $\mathbf{AAM}_{T}^{2}$ and $\mathbf{AAM}_{T}^{3}$, and force the $\mathbf{AAM}_{S}^{2}$ to learn from $\mathbf{AAM}_{T}^{2}$, $\mathbf{AAM}_{T}^{3}$ and $\mathbf{AAM}_{T}^{4}$. We denote $N_{T}$ as the number of transformer encoder layers for the teacher network. Mathematically, the objective function of the proposed ATARD can be expressed as:
\begin{equation}
\label{equ:loss function of ATARD}
\begin{aligned}
l_{ATARD} &= CE(S^{t}_{\theta }(\mathbf{x}^{adv}),y)+\sum_{k=1}^{N_{T}-1}\left \| \mathbf{AAM}_{T}^{k}-\mathbf{AAM}_{S}^{1} \right \|_{2}\\&+\sum_{k=2}^{N_{T}}\left \| \mathbf{AAM}_{T}^{k}-\mathbf{AAM}_{S}^{2} \right \|_{2},
\end{aligned}
\end{equation}
where $\mathbf{x}^{adv}$ is generated in each batch during each training iteration using the 3-step PGD attacks. The first term in \eqref{equ:loss function of ATARD}, $CE(S^{t}_{\theta }(\mathbf{x}^{adv}),y)$ is denoted as Loss1, and the rest is denoted as Loss2. The algorithm of the proposed ATARD is depicted in Algorithm \ref{alg:ATARD}. For clear comparison, a table containing the objective function of the ATARD and the related works is presented in Table \ref{tab:different onjective function}. As seen, the key contribution and the biggest difference between the proposed ATARD and the listed related works is that, instead of transferring the adversarial logits information, the ATARD transfers the adversarial attention information from the teacher model to the student model for better robustness against adversarial examples.

\begin{table*}[]
\begin{center}
\caption{The objective function of the proposed ATARD and the related works including AT, ARD, IAD, AKD, and RSLAD.}
\label{tab:different onjective function}
\scalebox{1.0}{
\begin{tabular}{|l|l|}

\hline
\textbf{Method} & \textbf{Objective Function}  \\ \hline
AT     &    $\alpha CE(f(\mathbf{x}),y)+(1-\alpha) CE(f(\mathbf{x}^{adv}),y)$       \\ \hline
ARD    &     $\alpha t^{2}KL(S^{t}_{\theta }(\mathbf{x}^{adv}), T^{t}(\mathbf{x}))+(1-\alpha )CE(S^{t}_{\theta }(\mathbf{x}),y)$         \\ \hline
IAD    &     $\alpha KL(S^{t}_{\theta }(\mathbf{x}^{adv}), T^{t}(\mathbf{x}))+(1-\alpha )KL(S^{t}_{\theta }(\mathbf{x}^{adv}), S^{t}(\mathbf{x}))$         \\ \hline
AKD    &   $(1-\lambda _{1}-\lambda _{2})\cdot  CE(S_{\theta }(\mathbf{x}^{adv}), y)+\lambda _{1}\cdot KL(S_{\theta }(\mathbf{x}^{adv}), T_{at}(\mathbf{x}))+\lambda _{2}\cdot KL(S_{\theta }(\mathbf{x}^{adv}), T_{std}(\mathbf{x}))$           \\ \hline
RSLAD   &  $\alpha KL(S^{t}_{\theta }(\mathbf{x}^{adv}), T(\mathbf{x}))+(1-\alpha )KL(S_{\theta }(\mathbf{x}), T(\mathbf{x}))$     \\ \hline
\textbf{ATARD} (proposed)   &    $CE(S^{t}_{\theta }(\mathbf{x}^{adv}),y)+\sum_{k=1}^{N_{T}-1}\left \| \mathbf{AAM}_{T}^{k}-\mathbf{AAM}_{S}^{1} \right \|_{2}+\sum_{k=2}^{N_{T}}\left \| \mathbf{AAM}_{T}^{k}-\mathbf{AAM}_{S}^{2} \right \|_{2}$              \\ \hline

\end{tabular}}
\end{center}
\end{table*}

\begin{algorithm}[ht!]
\caption{Attention-based Adversarial Robustness Distillation (ATARD)}\label{alg:ATARD}
\hspace*{\algorithmicindent}\textbf{Input: }
\begin{itemize}[leftmargin=1.1cm]
\item student model S, teacher model T, training dataset $D=\left \{ (\mathbf{x}_{i},y_{i}) \right \}_{i=1}^{n}$.
\item learning rate $\eta$, number of epochs $N$, batch size $m$, number of batches $M$.
\item adversarial attention map for teacher and student networks $\mathbf{AAM}_{T}$ and $\mathbf{AAM}_{S}$.
\item the number of the transformer encoder layers for teacher network $N_{T}$.
\end{itemize}
\hspace*{\algorithmicindent}\textbf{Output: }adversarially robust model $S_{r}$

\begin{algorithmic}[1]
\vspace{2mm}
\State \textbf{for} $epoch=1,...,N$ \textbf{do}
\State ~~~\textbf{for} mini-batch=$1,...,M$ \textbf{do}
\State ~~~~~~Sample a mini-batch $\left \{ (\mathbf{x}_{i},y_{i}) \right \}_{i=1}^{m}$ from $D$
\State ~~~~~~\textbf{for} $i=1,...,m$ (in parallel) \textbf{do}
\State ~~~~~~~~~~Obtain adversarial data $\mathbf{x}_{i}^{adv}$ of $\mathbf{x}_{i}$ by PGD attacks.
\State ~~~~~~\textbf{end for}
\State ~~~~~~~$\theta \leftarrow \theta -\eta \triangledown _{\theta }\{ CE(S^{t}_{\theta }(\mathbf{x}_{i}^{adv}),y_{i})+\sum_{k=1}^{N_{T}-1} \| \mathbf{AAM}_{T}^{k}-\mathbf{AAM}_{S}^{1} \|_{2}+\sum_{k=2}^{N_{T}}\|\mathbf{AAM}_{T}^{k}-\mathbf{AAM}_{S}^{2}\|_{2} \}$
\State ~~~\textbf{end for}
\State \textbf{end for} 
\end{algorithmic}
\end{algorithm}

\begin{figure}[ht]
\centering
\includegraphics[width=\columnwidth]{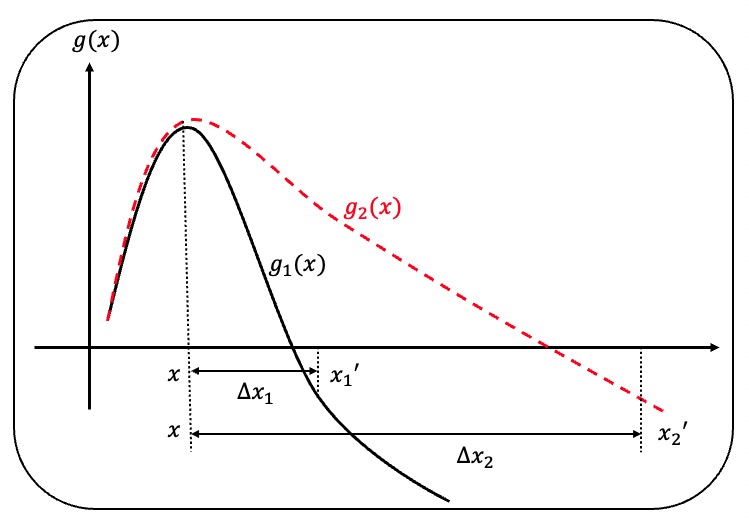}
  \caption{An illustrative example of the "smoothness" of a neural network.}
  \label{fig:smooth function}
\end{figure}

Now we shed light onto the reason as to why the proposed ATARD improves the robustness against adversarial examples through the perspective of the "smoothness" of the neural network. We first give a simple 2-D example to illustrate how a smoother function could help the robustness of a neural network against adversarial examples as shown in Figure \ref{fig:smooth function}. Given two neural networks $g_{1}(x)$ and $g_{2}(x)$ for two-class classification, we assume that the sample $x$ belongs to class 1 when $g(x)\geq  0$, otherwise, $x$ belongs to class 2. Starting from the same input $x$ which belongs to class 1, for $g_{1}(x)$, the perturbation $\Delta x_{1}$ needs to be added so that the perturbed sample $x_{1}'$ falls into the class 2, i.e., $x_{1}'$ is the adversarial version of $x$ for $g_{1}(x)$. Similarly, as for $g_{2}(x)$, the perturbation $\Delta x_{2}$ needs to be introduced in order for the perturbed sample $x_{2}'$ to become an adversarial sample. It is apparent that for a smoother function $g_{2}(x)$, it requires a larger perturbation for the sample $x$ to become an adversarial example. In other words, given the same amount of the perturbation, a smoother function could achieve higher robustness against adversarial examples. To verify the smoothness of our proposed ATARD, we calculate the average $l_{2}$-norm of the gradient of the loss function $CE(f(\mathbf{x}),y)$ with respect to the input for 1000 randomly chosen samples from the testing dataset. The results shown in Table \ref{tab:smoothness} demonstrate that the proposed ATARD has the smallest gradient norm among all the baseline methods, i.e., ATARD is the least sensitive to the adversarial perturbations among all the techniques. In this case, given a certain amount of perturbations, the ATARD could achieve higher robustness against adversarial examples as verified by the experimental results in section \uppercase\expandafter{\romannumeral4}, i.e., the attacker has to apply higher perturbations to make the input signals misclassified, hence more transmission power is needed to succeed with the modulation classification attacks, which will hinder stealth operation of the adversarial transmitter.

\begin{table}[h]
\caption{The average $l_{2}$-norm of the gradient of the loss function $CE(f(\mathbf{x}),y)$ with respect to the input.}
\label{tab:smoothness}
\begin{center}
\scalebox{1.0}{
\centering
\begin{tabular}{P{2.5cm} P{2.5cm}}
\hline
Methods   &  Gradient Norm  \\ \hline
NT             & 0.066  \\ \hline
AT             & 0.039  \\ \hline
ARD             & 0.031 \\ \hline
IAD             & 0.060 \\ \hline
AKD             & 0.040 \\ \hline
RSLAD             & 0.032 \\ \hline
\textbf{ATARD}             & \textbf{0.015} \\ \hline
\end{tabular}}
\end{center}
\end{table}

\subsection{White-box FGM and PGD attacks}
Now we give details of the white-box FGM and PGD attacks we used to evaluate the robustness of the proposed defense. According to the attacker's knowledge, adversarial examples can be classified into three different classes: perfect-knowledge white-box attacks, limited-knowledge gray-box attacks, and zero-knowledge black-box attacks. The white-box attack indicates the adversary has the full knowledge of the targeted defense system including the architectures, the parameters, and the data used by the defender. On the contrary, the black-box attack means the attacker has no knowledge of the targeted system, and the grey-box attack means the attacker has part of the knowledge of the defense. In this work, the white-box attack is considered, because this scenario allows for a worst-case evaluation of the security of the learning algorithms, creating empirical upper bounds on the performance deterioration that may be induced by the system under attack \cite{Biggio2018}.

The algorithm for generating white-box FGM attack is shown in Algorithm \ref{alg:FGM} which is adopted from \cite{sadeghi2018adversarial}. Specifically, given an input signal $\mathbf{x}_{0}$, certain PNR and SNR (i.e., the ratio of the signal power to the noise power), the allowed $l_{2}$-norm of the perturbation $\varepsilon$ is calculated as:
\begin{equation}
\label{equ:perturbation norm}
\varepsilon =\sqrt{\frac{\textup{PNR} \cdot  \left \| \mathbf{x}_{0} \right \|_{2}^{2}}{\textup{SNR}+1}}
\end{equation}
The above formula is obtained as described below. The dataset of RML (denote as $\mathbf{x}$) contains both the signal (assume the signal power is $S$) and noise (assume the noise power is $N$). The power of the data $\mathbf{x}$ (denoted by $P_{x}$) is therefore addition of the signal power and the noise power (because they are independent), i.e., $P_{x}=S+N$. Then we can obtain that: $\frac{P_{x}}{N}=\frac{S+N}{N}=\textup{SNR}+1$. Hence, $N=\frac{P_{x}}{\textup{SNR}+1}$. Given the $l_{2}$-norm of the adversarial perturbation $\varepsilon$, the perturbation power is $\varepsilon^{2}/L$, where $L$ is the number of elements in the perturbation vector. Then by definition, $\textup{PNR}=\frac{\varepsilon ^{2}/L}{N}=\varepsilon^{2}\cdot \frac{\textup{SNR}+1}{P_{x}L}$. Hence $\varepsilon =\sqrt{\frac{\textup{PNR}\cdot P_{x}L}{\textup{SNR}+1}}$ and \eqref{equ:perturbation norm} is obtained by replacing $P_{x}L$ with its sample estimate, i.e., $\left \| \mathbf{x}_{0} \right \|_{2}^{2}$. For each possible targeted wrong class, the normalized perturbation is obtained using line 3, where $\triangledown _{x_{0}}CE(f(\mathbf{x}_{0}), e_{t})$ indicates the gradient of the cross-entropy loss between the predicted probability $f(\mathbf{x}_{0})$ and the targeted wrong class $e_{t}$ with respect to the input $\mathbf{x}_{0}$. Then the normalized perturbation is multiplied by the allowed $l_{2}$-norm of the perturbation $\varepsilon$ and then added to the original sample in line 4. The algorithm will terminate when the predicted label of the generated adversarial sample $\argmax_{i}f_{i}(\mathbf{x}^\prime)$ is not equal to the true label $y$.

\begin{algorithm}[ht!]
\caption{FGM-based Adversarial Examples}\label{alg:FGM}
\hspace*{\algorithmicindent}\textbf{Input: }
\begin{itemize}[leftmargin=1.1cm]
\item input $\mathbf{x}_{0}$ and its true label $y$ 
\item the number of classes $N_{c}$
\item the prediction probability of the data sample $f(\cdot)$
\item allowed $l_{2}$-norm of the perturbation $\varepsilon$, allowed PNR and SNR
\item the cross entropy loss between the predicted probability and the targeted wrong class $CE(f(\cdot), e_{t})$

\end{itemize}
\hspace*{\algorithmicindent}\textbf{Output: }$\mathbf{x}^\prime$: the adversarial examples.

\begin{algorithmic}[1]
\State \textbf{for} $t$ in $range(N_{c})$ \textbf{do}
\State ~~~~$r_{norm}=- (\left \| \triangledown _{\mathbf{x}_{0}}CE(f(\mathbf{x}_{0}), e_{t}) \right \|_{2})^{-1}\triangledown _{\mathbf{x}_{0}}CE(f(\mathbf{x}_{0}), e_{t})$
\State ~~~$\mathbf{x}^\prime =\mathbf{x}_{0}+\varepsilon \cdot r_{norm}$

\State \textbf{until} $\argmax_{i}f_{i}(\mathbf{x}^\prime)\neq y$
\State \Return{$\mathbf{x}^\prime$}

\end{algorithmic}
\end{algorithm}

The PGD attack is considered as it is a strong form of attack utilizing the local first order information about the network \cite{madry2017towards}. The algorithm for generating the white-box PGD attack is shown in Algorithm \ref{alg:PGD}, which is adopted from \cite{madry2017towards}. Specifically, in line 4, the gradient of the cross-entropy loss between the predicted probability $f(\mathbf{x})$ and the true label $y$ is first computed and a standard gradient descent algorithm is employed to obtain the updated sample $\mathbf{x}^{*}$. Then a projection procedure $\Pi$ is applied to $\mathbf{x}^{*}$ in line 5 such that the generated adversarial perturbation is less than a predefined bound $\varepsilon$, which is obtained as in \eqref{equ:perturbation norm}. The projection procedure is formulated as the optimization below:
\begin{equation}
    \begin{aligned}
    \label{equ:projection1}
    &\min_{\mathbf{x}^\prime}\left \| \mathbf{x}^\prime - \mathbf{x}^{*} \right \|_{2}^{2},\\&
    s.t.~\left \| \mathbf{x}^\prime -\mathbf{x}_{0}\right \|_{2}\leq \varepsilon \\
    \end{aligned}
\end{equation}
where $\mathbf{x}^{*}$ indicates the updated sample after the standard gradient step $\mathbf{x}^{*} = \mathbf{x} + \eta_{0} \triangledown CE(f(\mathbf{x}), y)$ and $\mathbf{x}_{0}$ is the original input signal.
The solution to \eqref{equ:projection1} is computed as following:
\begin{equation}
    \label{equ:projection11}
     \mathbf{x}^\prime = \mathbf{x}_{0}+\frac{\mathbf{x}^{*}-\mathbf{x}_{0}}{max(\varepsilon, \left \| \mathbf{x}^{*}-\mathbf{x}_{0} \right \|_{2})}\cdot \varepsilon.   
\end{equation}
However, we adopt \eqref{equ:projector} as the projector in order to make the $l_{2}$-norm of the generated adversarial perturbation equal to $\varepsilon $, i.e., $\left \| \mathbf{x}^\prime-\mathbf{x}_{0} \right \|_{2}=\varepsilon$, which will help setting specific PNR for performance analysis.

\begin{equation}
\label{equ:projector}
\mathbf{x}^\prime=\mathbf{x}_{0}+\frac{\varepsilon \cdot (\mathbf{x}^{*}-\mathbf{x}_{0})}{\left \| \mathbf{x}^{*}-\mathbf{x}_{0} \right \|_{2}}
\end{equation}
Finally, the loop will terminate when the condition in line 5 is met, i.e., the predicted label of the produced adversarial example $\argmax_{i}f_{i}(\mathbf{x}^\prime)$ is unequal to the true label $y$.

\begin{algorithm}[ht!]
\caption{PGD-based Adversarial Examples}\label{alg:PGD}
\hspace*{\algorithmicindent}\textbf{Input: }
\begin{itemize}[leftmargin=1.1cm]
\item input $\mathbf{x}_{0}$ and its true label $y$
\item the step size $\eta_{0}$
\item the prediction probability of the data sample $f(\cdot)$
\item the cross entropy loss between the predicted probability and the true label $CE(f(\cdot), y)$
\item a projector $\Pi$ on the $l_{2}$-norm constraint $|| \mathbf{x}^\prime-\mathbf{x}_{0}||_{2}\leq \varepsilon$, where $\varepsilon =\sqrt{\frac{\textup{PNR} \cdot  \left \| \mathbf{x}_{0} \right \|_{2}^{2}}{\textup{SNR}+1}}$
\end{itemize}
\hspace*{\algorithmicindent}\textbf{Output: }$\mathbf{x}^\prime$: the adversarial examples.

\begin{algorithmic}[1]

\State $\mathbf{x}^\prime\gets \mathbf{x}_{0}$

\Repeat
\State $\mathbf{x}\gets \mathbf{x}^\prime$
\State $\mathbf{x}^{*}\gets \mathbf{x} + \eta_{0} \triangledown CE(f(\mathbf{x}), y)$
\State $\mathbf{x}^\prime\gets \Pi (\mathbf{x}^{*})$

\Until $\argmax_{i}f_{i}(\mathbf{x}^\prime)\neq y$

\State \Return{$\mathbf{x}^\prime$}

\end{algorithmic}
\end{algorithm}

\section{Results and Discussion}
In this section, we present and analyze the experimental results. All algorithms are implemented in PyTorch and executed by NVIDIA GEforce RTX 2080 Ti GPU.
\subsection{Dataset}
The dataset used in this work is the GNU radio ML dataset RML2016.19a \cite{OShea2016} and RDL2021.12 \cite{luan2022automatic}. The GNU radio ML dataset RML2016.10a has 220,000 input samples, each of which corresponds to one modulation type at a specific SNR. This dataset contains 11 different modulation schemes including BPSK, QPSK, 8PSK, QAM16, QAM64, CPFSK, GFSK, PAM4, WBFM, AM-SSB, and AM-DSB. The samples are produced for 20 different SNR levels ranging from -20 dB to 18 dB in 2 dB steps. Each sample contains 256 dimensions, including 128 in-phase and 128 quadrature components. Half of the samples are used for training, while the rest is used for testing. Compared to RML dataset, the RDL dataset contains two noise types namely Gaussian noise and Alpha-stable distributed noise. In addition, the RDL dataset considered both the Rayleigh fading and Rician fading channels. 110,000 samples are generated for 11 different modulation types with ${\textup{SNR~=~10~dB}}$, in which 90$\%$ samples are used for training, and the rest are used for testing. The RDL dataset has impulsive noise due to addition of alpha-stable noise. Therefore, to mitigate the effect of impulsive noise, we have pre-processed the received signal before applying it to the DNN classifiers during the training and testing phases. Accordingly, we computed the standard deviation $\sigma_{x}$ of the signal after applying a five sample moving median filter and rejected any samples that fall outside of $\pm \sigma_{x}$. Finally the data samples were normalized. To assess the effectiveness of the transformer-based neural network against adversarial samples, we generate adversarial attacks using 1000 data samples from the testing set with $\textup{SNR~=~10~dB}$. 
\subsection{Robustness results against white-box FGM and PGD attacks}
\begin{figure}[ht!]
\centering
\includegraphics[width=\columnwidth]{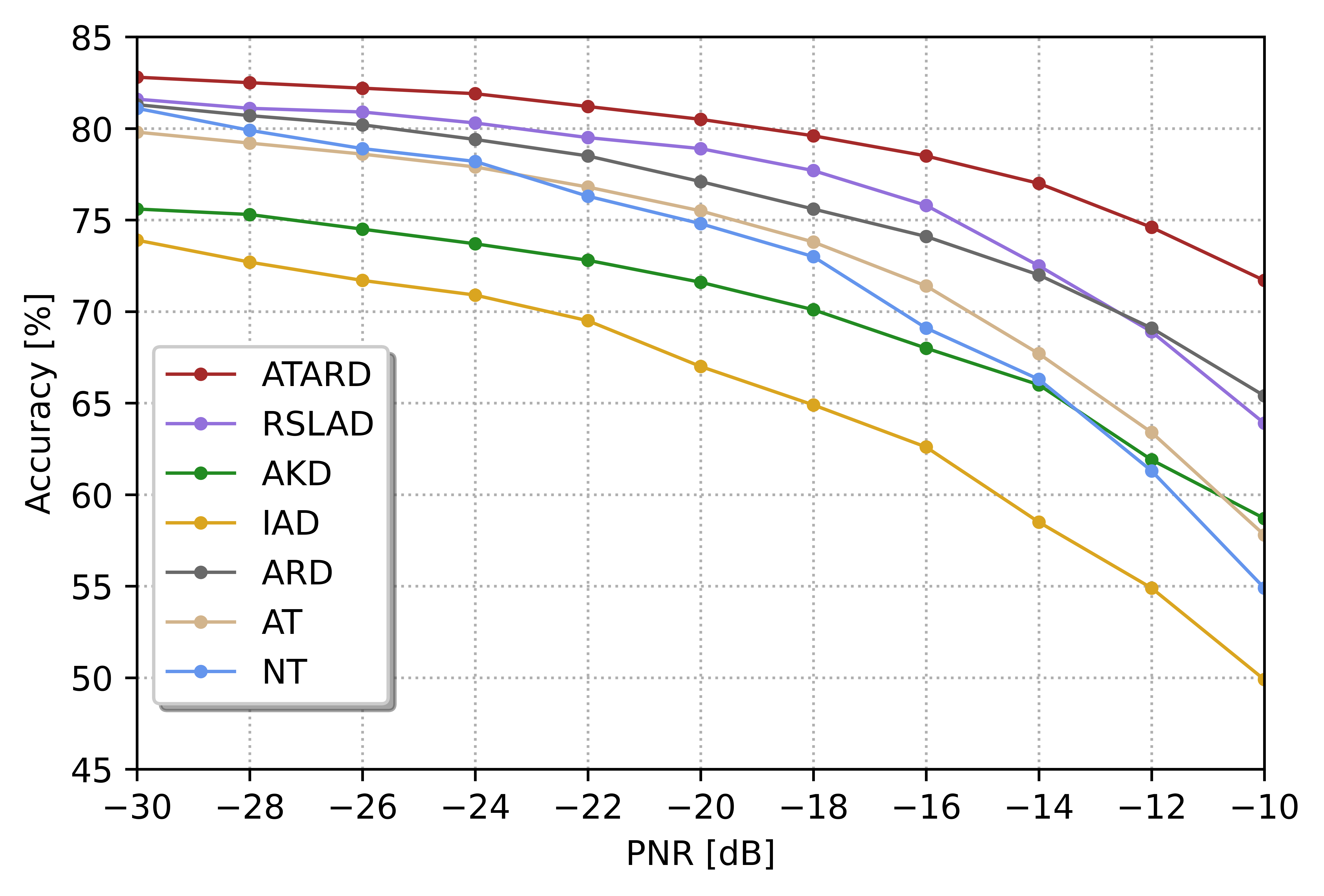}
  \caption{Classification accuracy against FGM attacks for a wide range of PNR values using RML dataset.}
  \label{fig:white-box_FGM}
\end{figure}

\begin{figure}[ht!]
\centering
\includegraphics[width=\columnwidth]{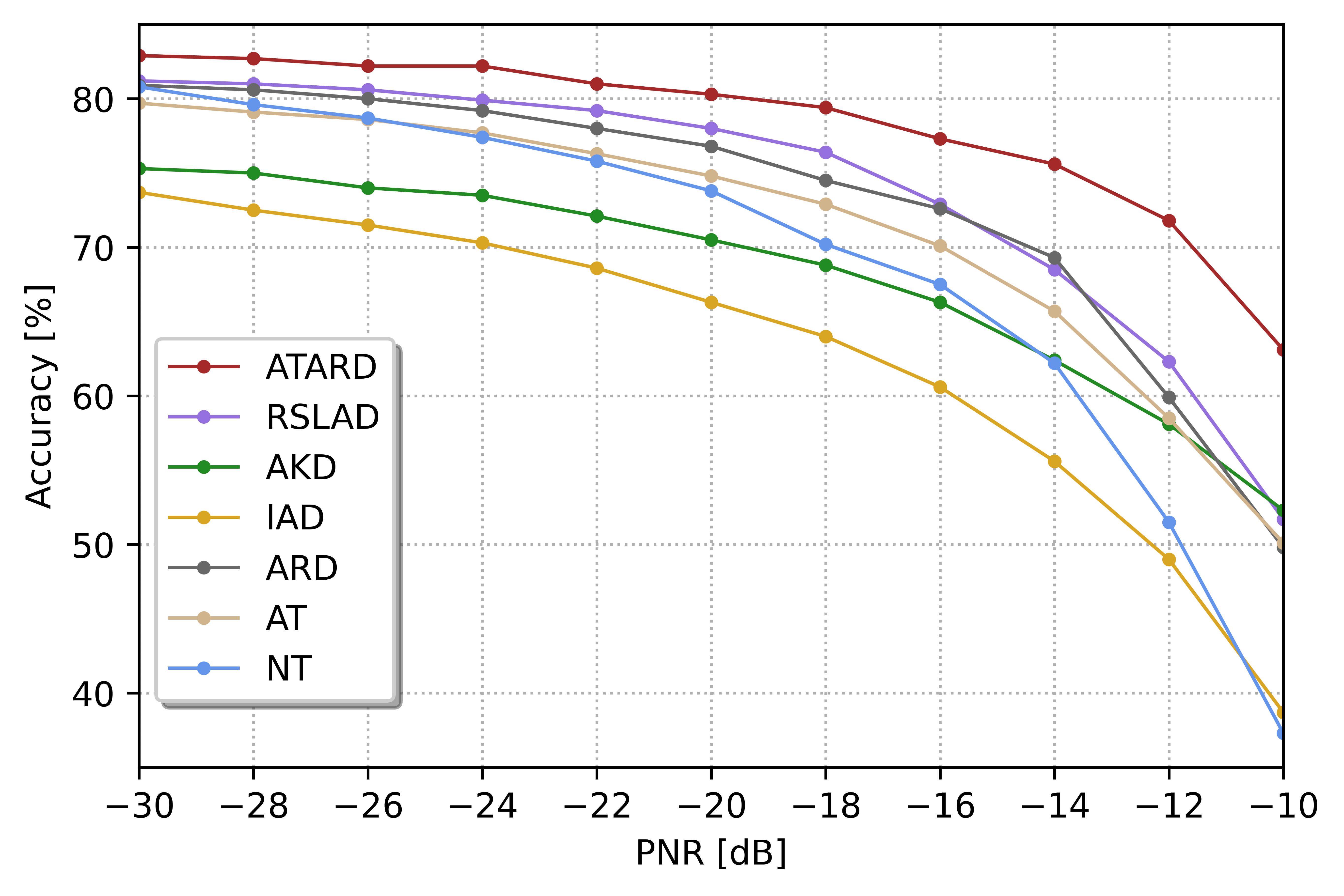}
  \caption{Classification accuracy against PGD attacks for a wide range of PNR values using RML dataset.}
  \label{fig:white-box_PGD}
\end{figure}

Considering RML2016.19a dataset, the classification accuracy of ATARD and other competing techniques in the presence of FGM and PGD attacks is shown in Figure \ref{fig:white-box_FGM} and Figure \ref{fig:white-box_PGD}, respectively. The experiments are repeated for ten times and the average performance is presented. Our proposed ATARD has higher robustness than the competing techniques including AT, ARD, IAD, AKD, and RSLAD for a wide range of PNR values from -30 dB to -10 dB. The results for normal training (NT)-based transformer are also included as the baseline. The robustness improvement becomes more significant when PNR value is large. Specifically, from Figure \ref{fig:white-box_FGM}, when $\textup{PNR}=-10~\textup{dB}$, the proposed ATARD achieves 71.7$\%$ accuracy which is around 7$\%$ higher than ARD, RSLAD and 14$\%$ higher than AT. From Figure \ref{fig:white-box_PGD}, when $\textup{PNR}=-10~\textup{dB}$, the classification accuracy against PGD attacks for the proposed ATARD is around 13$\%$ higher than AT, and around 10$\%$ higher than ARD, AKD, and RSLAD. As compared to the NT, the proposed ATARD improves the robustness against both FGM and PGD attacks significantly. Specifically, when $\textup{PNR}=-10~\textup{dB}$, the proposed ATARD achieves 16.8$\%$ and 25.8$\%$ higher accuracy against FGM and PGD attacks, respectively. Furthermore, from the attacker's point of view, the PGD attack is more powerful than the FGM attacks, i.e., the PGD attacks achieves higher misclassification rate than that of the FGM attacks for a wide range of PNR values from -30 dB to -10 dB. For the proposed ATARD, the misclassification rate against PGD attacks is 8.5$\%$ higher than that against FGM attacks when $\textup{PNR}=-10~\textup{dB}$.

\begin{figure}[ht!]
\centering
\includegraphics[width=\columnwidth]{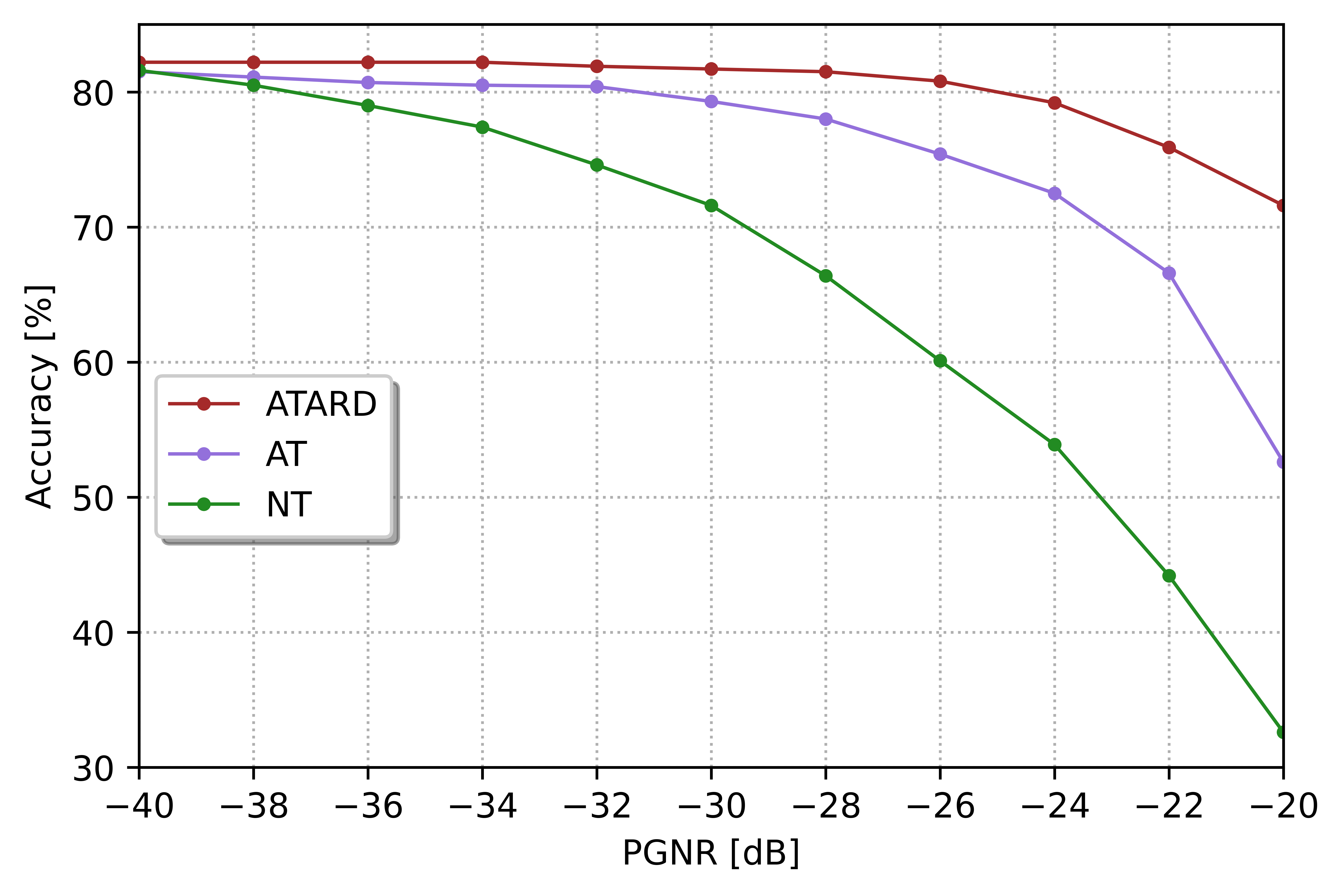}
  \caption{Classification accuracy against PGD attacks for a wide range of PGNR values using RDL dataset under a Rayleigh fading channel scenario.}
  \label{fig:Rayleigh_PGD}
\end{figure}

\begin{figure}[ht!]
\centering
\includegraphics[width=\columnwidth]{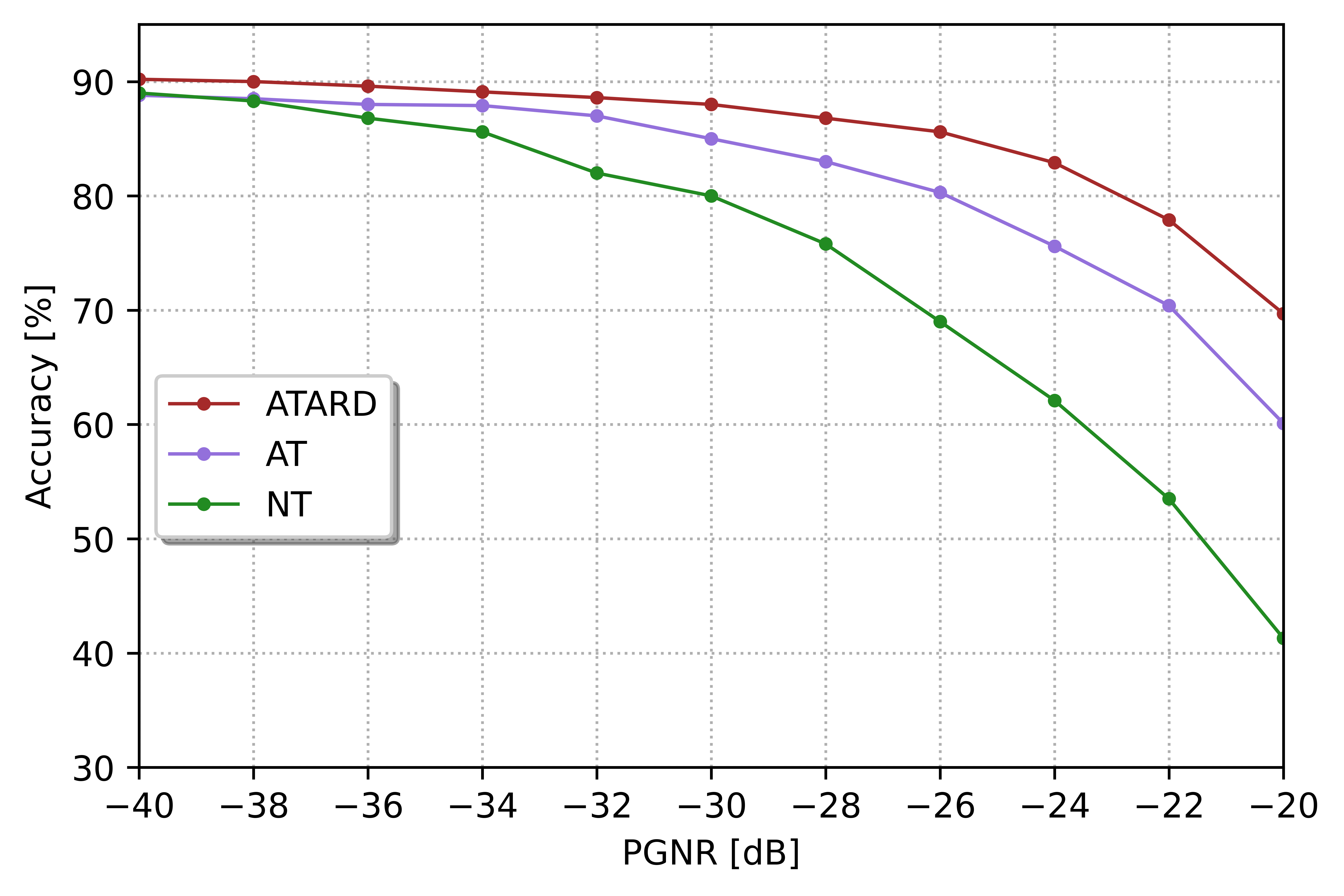} 
  \caption{Classification accuracy against PGD attacks for a wide range of PGNR values using RDL dataset under a Rician fading channel scenario.}
  \label{fig:Rician_PGD}
\end{figure}

Furthermore, considering RDL 2021.12 dataset, the classification accuracy against PGD attacks for a wide range of perturbation to generalized noise ratio (PGNR) values for both the Rayleigh and Rician fading channels is presented in Figure \ref{fig:Rayleigh_PGD} and Figure \ref{fig:Rician_PGD}, respectively. As in \cite{luan2022automatic}, we considered the identical power for the Gaussian noise and alpha-stable noise. Our definition of PGNR is the ratio of the perturbation power and the alpha-stable noise power. Therefore, the ratio of the perturbation to Gaussian noise power is same as PGNR, but the ratio of perturbation power to total Gaussian and alpha-stable noise power is $\textup{PGNR}/2$ (i.e 3dB less than PGNR in dB). Please note that for a given SNR and PGNR value, the perturbation to signal ratio (PSR) can be obtained as $\textup{PSR}=\textup{PGNR} / \textup{SNR}$. As shown in Figure \ref{fig:Rayleigh_PGD}, for Rayleigh fading channels, our proposed ATARD scheme which transfers the attention map outperforms the AT technique and standard training. The improvement of accuracy becomes more significant when PGNR is large. Specifically, when $\textup{PGNR}=-20~\textup{dB}$, the proposed ATARD achieves 19.0$\%$ higher accuracy than AT and 39.0$\%$ higher accuracy than NT. A similar trend is seen for Rician channel as shown in Figure \ref{fig:Rician_PGD}. Specifically, the proposed ATARD achieves 9.6$\%$ higher accuracy than AT and 28.4$\%$ higher accuracy than NT when $\textup{PGNR}=-20~\textup{dB}$.

\subsection{Transferability between different architectures}
In this section, we investigate the transferability of the proposed ATARD method. Transferability is a common phenomenon in adversarial examples. It is discovered in  \cite{papernot2016transferability} that the adversarial examples produced against a neural network can fool other neural networks with different architectures. Transferability is important for black-box attacks in which the target DL model and the training data set are unavailable. In this case, a surrogate neural network model can be trained by the attacker and adversarial examples can be produced against a surrogate model. Then, the target model could be vulnerable to these adversarial examples because of the transferability. In this work, to investigate the transferability between different architectures, we adopt two surrogate transformer architectures, and the settings of these surrogate transformer architectures are shown in Table \ref{tab:CNN_parameter}. The difference between the student transformer architecture and the two surrogate transformer architectures lies in different patch size, different embedding dimension, different number of transformer encoder layer and different number of hidden layer features as detailed in Table \ref{tab:CNN_parameter}. The total number of parameters for substitute Model-1 and Model-2 are 102,603 and 801,675, respectively. 
\FloatBarrier
\begin{table}[h] 
\caption{Settings of two substitute transformer architectures.}
\label{tab:CNN_parameter}
\begin{center}
\begin{tabular}{l l l}
\hline
\textbf{Settings}  & \textbf{ Model-1}   & \textbf{Model-2}                      \\ \hline
Patch size          &(2,16)   &(2,32)\\ 
Embedding dimension               &64  &128\\ 
number of transformer encoder layer                &2   &4                               \\ 
number of multi-head             &4  &4\\ 
number of hidden layer features                 & 256  &512\\                                                    \hline

\end{tabular}
\end{center}
\end{table}

\begin{figure}[ht!]
\centering
\includegraphics[width=\columnwidth]{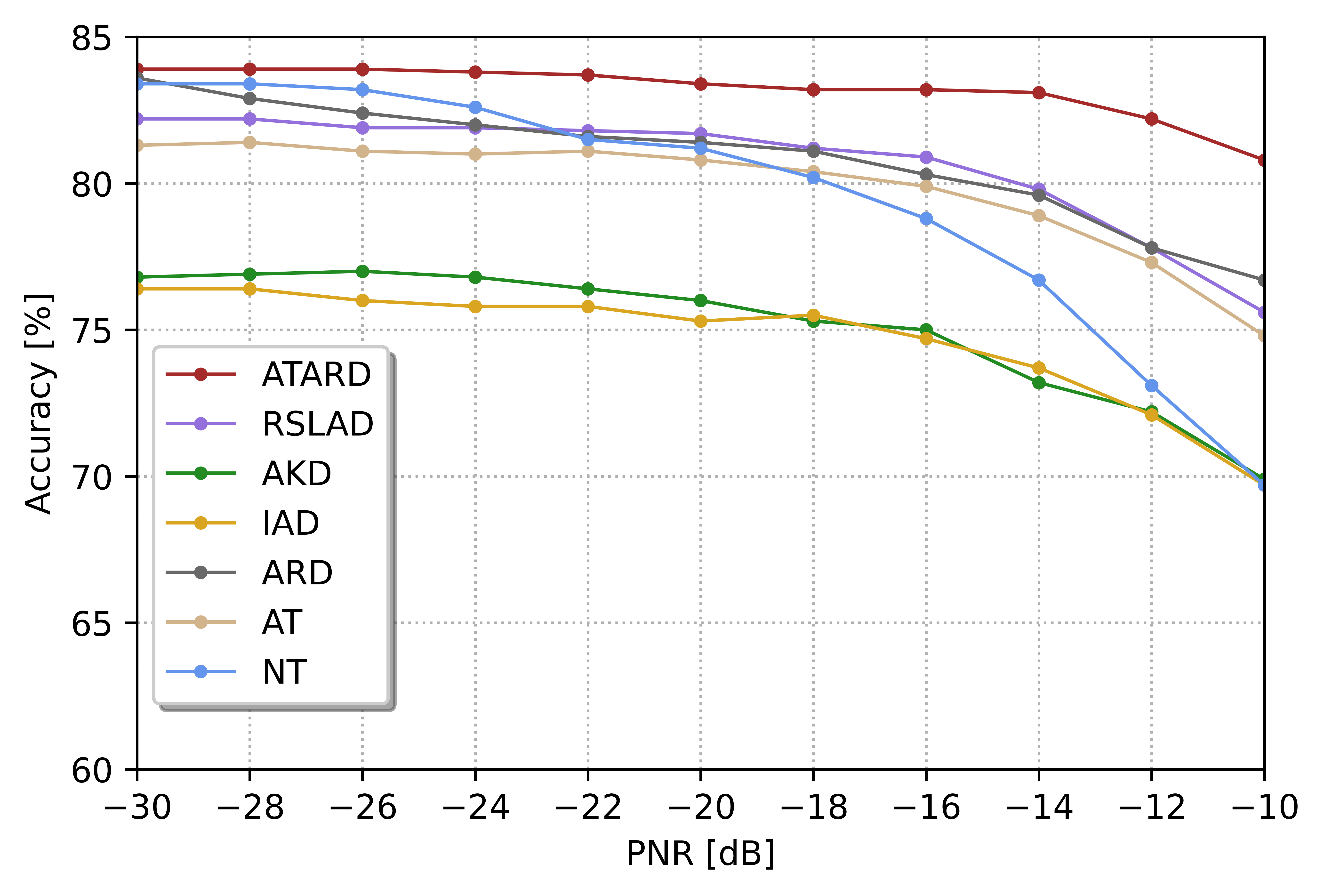}
  \caption{Classification accuracy against FGM attacks produced on Model-1 for a range of PNR values.}
  \label{fig:black-box_FGM_2mini}
\end{figure}

\begin{figure}[ht!]
\centering
\includegraphics[width=\columnwidth]{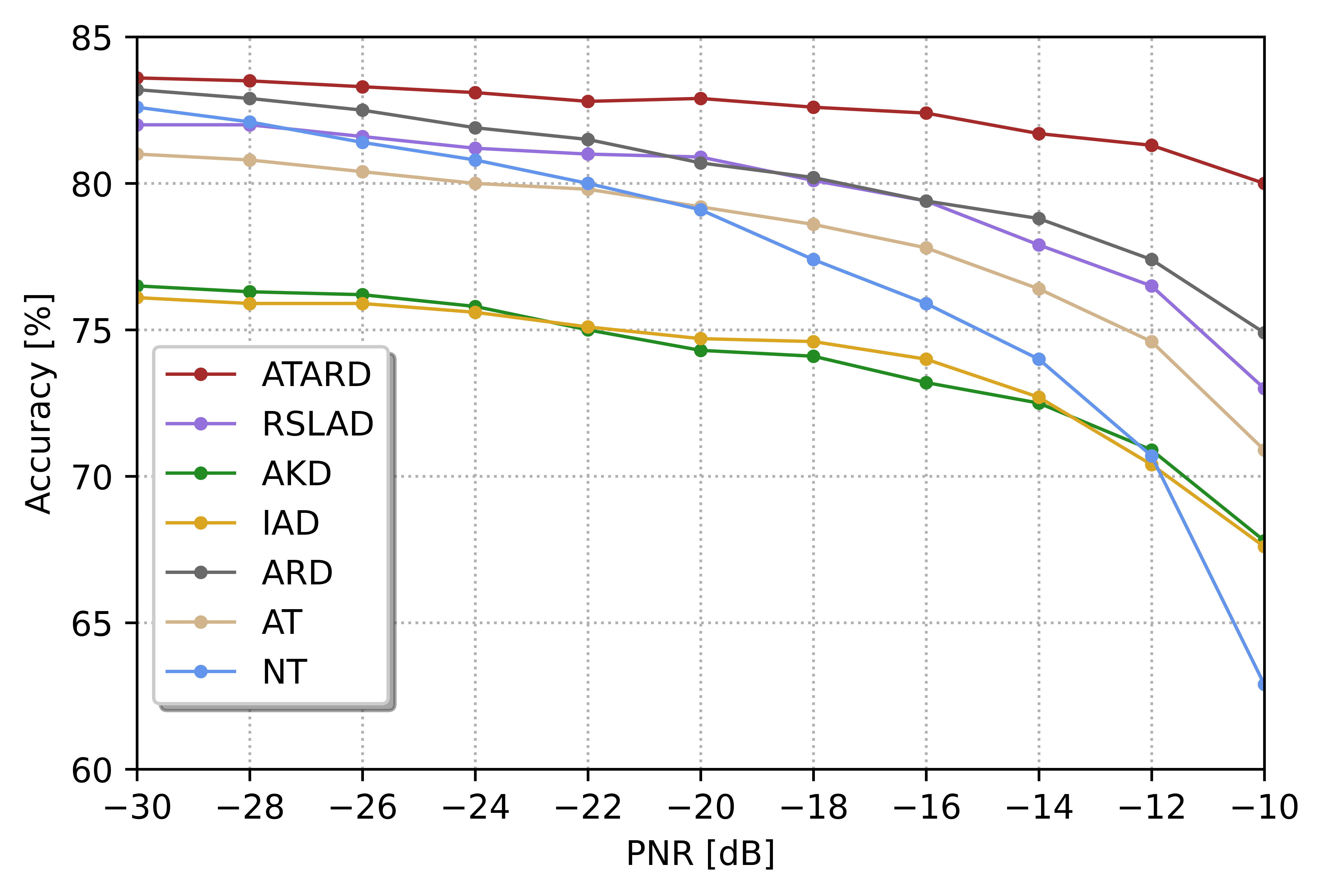}
  \caption{Classification accuracy against PGD attacks produced on Model-1 for a range of PNR values.}
  \label{fig:black-box_PGD_2mini}
\end{figure}

\begin{figure}[ht!]
\centering
\includegraphics[width=\columnwidth]{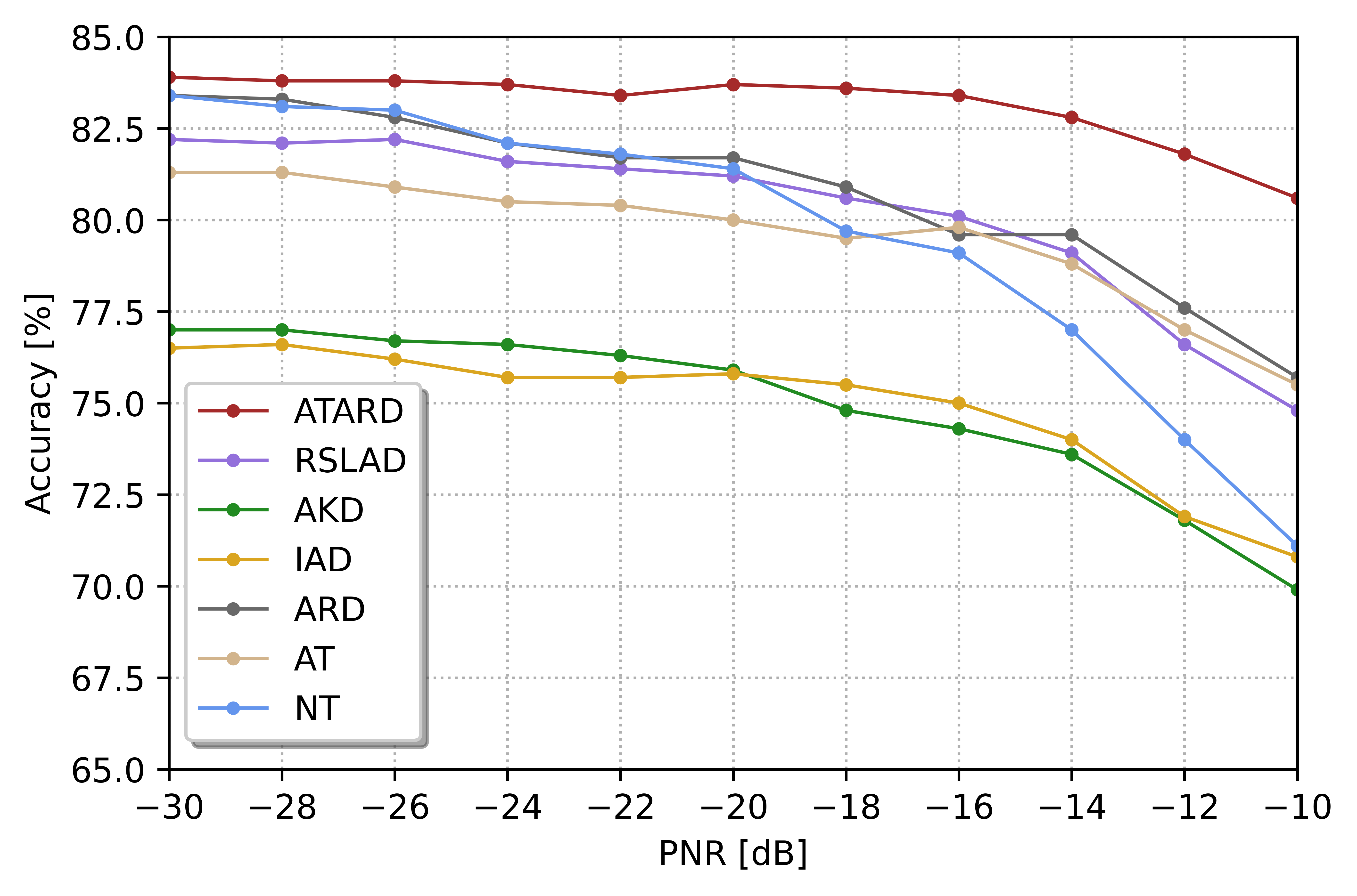}
  \caption{Classification accuracy against FGM attacks produced on Model-2 for a range of PNR values.}
  \label{fig:black-box_FGM_2big}
\end{figure}

\begin{figure}[ht!]
\centering
\includegraphics[width=\columnwidth]{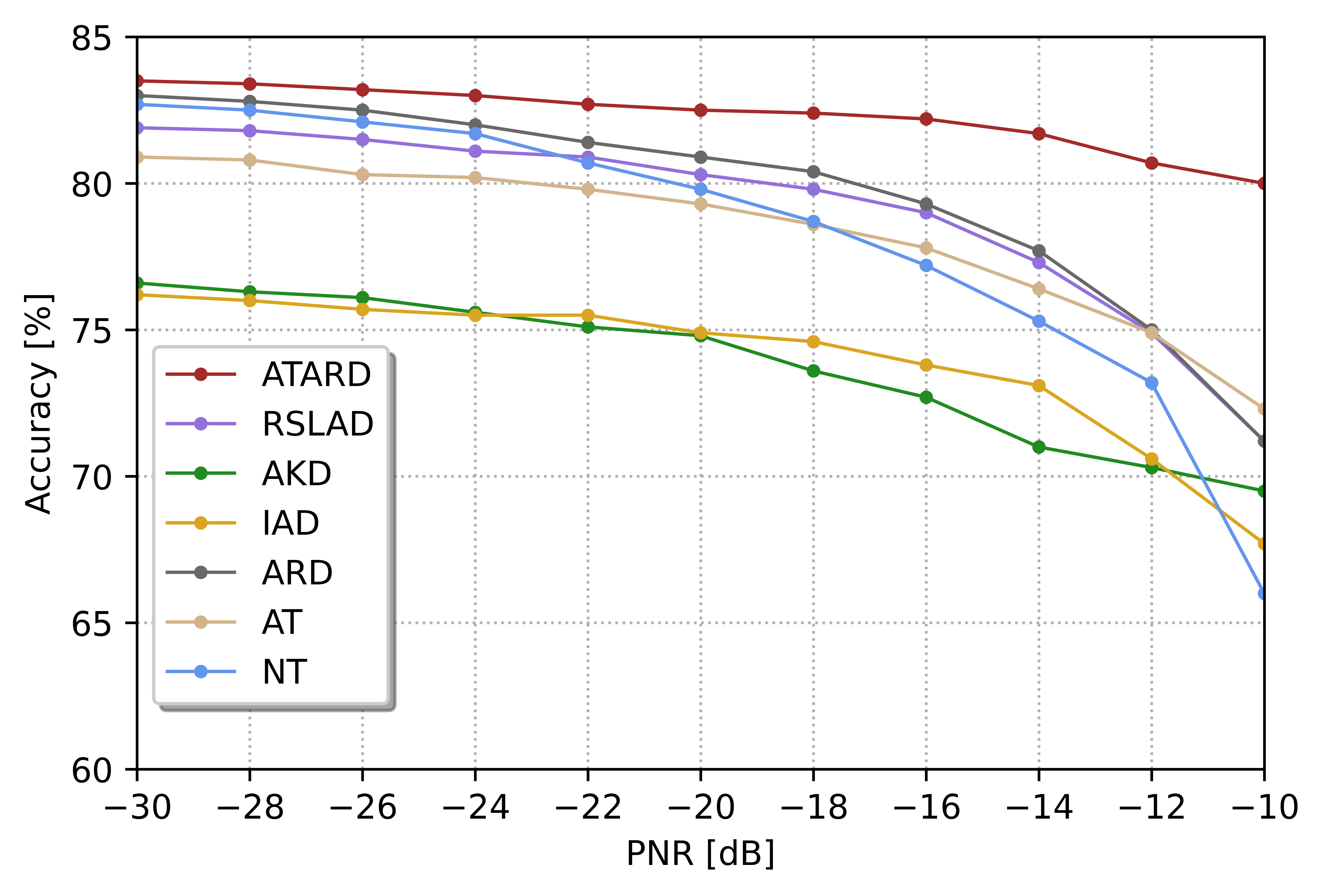}
  \caption{Classification accuracy against PGD attacks produced on Model-2 for a range of PNR values.}
  \label{fig:black-box_PGD_2big}
\end{figure}

To investigate the transferability of the proposed method, we use 1000 RML data samples from the testing dataset to generate FGM and PGD attacks on two substitute networks: Model-1 and Model-2. Then the produced adversarial attacks are applied to the trained ATARD-based transformer network to calculate the classification accuracy. The results are shown in Figure \ref{fig:black-box_FGM_2mini} - Figure \ref{fig:black-box_PGD_2big}. As expected, the adversarial examples maintain good transferability among different architectures. The PGD attacks generated based on both Model-1 and Model-2 could achieve around 36$\%$ misclassification rate (i.e., around 64$\%$ classification accuracy) when $\textup{PNR}=-10~\textup{dB}$ in the absence of a defense mechanism. From Figure \ref{fig:black-box_PGD_2mini}, it can be seen that with the defense system like the proposed ATARD, the transformer model can achieve a better classification accuracy against PGD attacks transferred from Model-1 for a wide range of PNR values. Specifically, the ATARD achieves 80.0$\%$ and 82.9$\%$ accuracy when $\textup{PNR}=-10~\textup{dB}$ and when $\textup{PNR}=-20~\textup{dB}$, respectively, which is roughly 17.1$\%$ and 3.8$\%$ higher than the NT trained transformer (i.e., in the absence of the defense mechanism). Similarly, in Figure \ref{fig:black-box_FGM_2mini}, the transformer in the presence of ATARD defense system performs better than the NT trained transformer from -30 dB to -10 dB. Similarly, for surrogate Model-2, it can be seen from Figure \ref{fig:black-box_FGM_2big} and Figure \ref{fig:black-box_PGD_2big} that with the defense system like the proposed ATARD, the transformer model can achieve a better classification accuracy against FGM and PGD attacks transferred from Model-2 for a wide range of PNR values. Specifically, using PGD attacks, the ATARD achieves 14.0$\%$ and 3$\%$ higher accuracy than the NT trained transformer when $\textup{PNR}=-10~\textup{dB}$ and $\textup{PNR}=-20~\textup{dB}$, respectively; In the mean time, using FGM attacks, the ATARD achieves 9.5$\%$ and 2.3$\%$ higher accuracy than the NT trained transformer when $\textup{PNR}=-10~\textup{dB}$ and $\textup{PNR}=-20~\textup{dB}$, respectively.

\section{Conclusion}

We have proposed for the first time in the literature  a transformer-based defense mechanism, namely an ATARD method in modulation classifications for low-power IoT devices. The main novelty is that the adversarial attention map was transferred from a large transformer network to a small transformer model. Through experimental results, we have shown that the proposed ATARD scheme achieves a better robustness against adversarial examples among all the related works including NT, AT, ARD, IAD, AKD, and RSLAD. We have also provided a possible reasoning for the superior performance of ATARD by investigating the smoothness of the loss function, i.e., the ATARD-based transformer model is less sensitive to the perturbations. As a result, the attacker will have to apply more transmission power to fool the ATARD-based transformer model used by the defender. The transferability of the adversarial examples among different architectures was also investigated to show the superior performance of the proposed ATARD-based transformer model against the transferability of adversarial examples.

\section*{Acknowledgments}
This work was supported in part by the U.K. Engineering
and Physical Sciences Research Council (EPSRC) under Grant EP/R006385/1
and Grant EP/N007840/1; in part by the International Scientific Partnership
Programme of King Saud University under Grant ISPP-18-134(2); and
in part by the CELTIC-NEXT European Collaborative Project AIMM
(ID: C2019/2-5) through the UKRI Innovate U.K. under Grant 48160. For the purpose of open access, the author(s) has applied a Creative Commons Attribution (CC BY) licence to any Author Accepted Manuscript version arising

\appendix
\subsection{Calculation of the number of parameters for the transformer network}
We compute the number of parameters used in our teacher and student transformer networks. The transformer can be divided into five parts including a convolutional layer, a CLS token, a stack of transformer encoder layers, an LN layer, and a dense layer. The number of the parameters for each part is calculated as below.

For a convolutional layer, given input channel $l$, output channel $k$, and the filter size $n*m$, its number of parameters $N_{\textup{CL}}$ can be calculated as:
\begin{equation}
\label{equ:N_CL}
N_{\textup{CL}}=(n*m*l+1)*k
\end{equation}
For a CLS token, as mentioned before, it has the same dimension as each embedded patch. Therefore, its number of parameters $N_{\textup{cls}}$ is equal to the output channel $k$, i.e., $N_{\textup{cls}}=k$. For a stack of N transformer encoder layers, its calculation can be divided into three parts including an MSA, two LN layers and an FFN. For the SA, given an input sequence $\mathbf{Z}$, three projections are used for queries $\mathbf{Q}$, keys $\mathbf{K}$ and values $\mathbf{V}$ as shown below:
\begin{equation}
\label{equ:Q}
\mathbf{Q}=\mathbf{Z}\mathbf{U_{Q}}
\end{equation}
\begin{equation}
\label{equ:K}
\mathbf{Q}=\mathbf{Z}\mathbf{U_{K}}
\end{equation}
\begin{equation}
\label{equ:V}
\mathbf{V}=\mathbf{Z}\mathbf{U_{V}}
\end{equation}
where $\mathbf{U_{Q}}$, $\mathbf{U_{K}}$ and $\mathbf{U_{V}}$ are three projection matrix. For the MSA, a linear projection matrix $\mathbf{U}_\textup{{MSA}}$ is used such that the outputs of SA operations can be concatenated and projected as shown in \eqref{equ:msa}. Hence the number of parameters for the MSA $N_{\textup{MSA}}$ is obtained as:
\begin{equation}
\label{equ:N_MSA}
N_{\textup{{MSA}}}=(k*k)*4+k
\end{equation}
For the LN layer, its number of parameters $N_{\textup{LN}}$ is obtained as:
\begin{equation}
\label{equ:N_LN}
N_{\textup{LN}}=2*k
\end{equation}
The FFN is a fully connected network with one hidden layer. Considering the size of the hidden layer $s$, its number of parameters $N_{\textup{FFN}}$ is written as:
\begin{equation}
\label{equ:N_FFN}
N_{\textup{FFN}}=2*k*s+k+s
\end{equation}
For a simple dense layer with c classes output, its number of parameters $N_{\textup{DL}}$ is calculated as:
\begin{equation}
\label{equ:N_DL}
N_{\textup{DL}}=k*c+c
\end{equation}
Finally, for the transformer network, the total number of the parameters is:
\begin{equation}
\label{equ:N_total}
\begin{aligned}
N_{total}&=N_{\textup{CL}} + N_{\textup{cls}}+N_{\textup{LN}}+N_{\textup{DL}}\\&+N*(N_{\textup{{MSA}}}+2*N_{\textup{LN}}+N_{\textup{FFN}})
\end{aligned}
\end{equation}
For the teacher transformer network, we have $n$=2, $m$=32, $l$=1, $k$=128, $N$=4, $s$=512, $c$=11; For the student network, we have $n$=2, $m$=32, $l$=1, $k$=96, $N$=2, $s$=384, $c$=11. Hence, the number of parameters for the teacher and student network is 801,675 and 230,699, respectively.

\small{
    \bibliographystyle{IEEEtran}
    \bibliography{references}

\begin{thebibliography}{10}
\providecommand{\url}[1]{#1}
\csname url@samestyle\endcsname
\providecommand{\newblock}{\relax}
\providecommand{\bibinfo}[2]{#2}
\providecommand{\BIBentrySTDinterwordspacing}{\spaceskip=0pt\relax}
\providecommand{\BIBentryALTinterwordstretchfactor}{4}
\providecommand{\BIBentryALTinterwordspacing}{\spaceskip=\fontdimen2\font plus
\BIBentryALTinterwordstretchfactor\fontdimen3\font minus
  \fontdimen4\font\relax}
\providecommand{\BIBforeignlanguage}[2]{{%
\expandafter\ifx\csname l@#1\endcsname\relax
\typeout{** WARNING: IEEEtran.bst: No hyphenation pattern has been}%
\typeout{** loaded for the language `#1'. Using the pattern for}%
\typeout{** the default language instead.}%
\else
\language=\csname l@#1\endcsname
\fi
#2}}
\providecommand{\BIBdecl}{\relax}
\BIBdecl

\bibitem{cui2021integrating}
Y.~Cui, F.~Liu, X.~Jing, and J.~Mu, ``Integrating sensing and communications
  for ubiquitous iot: Applications, trends, and challenges,'' \emph{IEEE
  Network}, vol.~35, no.~5, pp. 158--167, 2021.

\bibitem{mu2021machine}
J.~Mu, X.~Jing, Y.~Zhang, Y.~Gong, R.~Zhang, and F.~Zhang, ``Machine
  learning-based 5g ran slicing for broadcasting services,'' \emph{IEEE
  Transactions on Broadcasting}, 2021.

\bibitem{zhang2020device}
R.~Zhang, X.~Jing, S.~Wu, C.~Jiang, J.~Mu, and F.~R. Yu, ``Device-free wireless
  sensing for human detection: the deep learning perspective,'' \emph{IEEE
  Internet of Things Journal}, vol.~8, no.~4, pp. 2517--2539, 2020.

\bibitem{wang2019energy}
Z.~Wang, R.~Liu, Q.~Liu, J.~S. Thompson, and M.~Kadoch, ``Energy-efficient data
  collection and device positioning in uav-assisted iot,'' \emph{IEEE Internet
  of Things Journal}, vol.~7, no.~2, pp. 1122--1139, 2019.

\bibitem{weber2015automatic}
C.~Weber, M.~Peter, and T.~Felhauer, ``Automatic modulation classification
  technique for radio monitoring,'' \emph{Electronics Letters}, vol.~51,
  no.~10, pp. 794--796, 2015.

\bibitem{clancy2007applications}
C.~Clancy, J.~Hecker, E.~Stuntebeck, and T.~O'Shea, ``Applications of machine
  learning to cognitive radio networks,'' \emph{IEEE Wireless Communications},
  vol.~14, no.~4, pp. 47--52, 2007.

\bibitem{wu2008novel}
H.-C. Wu, M.~Saquib, and Z.~Yun, ``Novel automatic modulation classification
  using cumulant features for communications via multipath channels,''
  \emph{IEEE Transactions on Wireless Communications}, vol.~7, no.~8, pp.
  3098--3105, 2008.

\bibitem{ramkumar2009automatic}
B.~Ramkumar, ``Automatic modulation classification for cognitive radios using
  cyclic feature detection,'' \emph{IEEE Circuits and Systems Magazine},
  vol.~9, no.~2, pp. 27--45, 2009.

\bibitem{park2008automatic}
C.-S. Park, J.-H. Choi, S.-P. Nah, W.~Jang, and D.~Y. Kim, ``Automatic
  modulation recognition of digital signals using wavelet features and svm,''
  in \emph{2008 10th International Conference on Advanced Communication
  Technology}.\hskip 1em plus 0.5em minus 0.4em\relax IEEE, 2008, pp. 387--390.

\bibitem{swami2000hierarchical}
A.~Swami and B.~M. Sadler, ``Hierarchical digital modulation classification
  using cumulants,'' \emph{IEEE Transactions on communications}, vol.~48,
  no.~3, pp. 416--429, 2000.

\bibitem{soliman1992signal}
S.~S. Soliman and S.-Z. Hsue, ``Signal classification using statistical
  moments,'' \emph{IEEE Transactions on Communications}, vol.~40, no.~5, pp.
  908--916, 1992.

\bibitem{o2016convolutional}
T.~J. O’Shea, J.~Corgan, and T.~C. Clancy, ``Convolutional radio modulation
  recognition networks,'' in \emph{Engineering Applications of Neural Networks:
  17th International Conference, EANN 2016, Aberdeen, UK, September 2-5, 2016,
  Proceedings 17}.\hskip 1em plus 0.5em minus 0.4em\relax Springer, 2016, pp.
  213--226.

\bibitem{o2018over}
T.~J. O’Shea, T.~Roy, and T.~C. Clancy, ``Over-the-air deep learning based
  radio signal classification,'' \emph{IEEE Journal of Selected Topics in
  Signal Processing}, vol.~12, no.~1, pp. 168--179, 2018.

\bibitem{krzyston2020high}
J.~Krzyston, R.~Bhattacharjea, and A.~Stark, ``High-capacity complex
  convolutional neural networks for {I/Q} modulation classification,''
  \emph{arXiv preprint arXiv:2010.10717}, 2020.

\bibitem{liao2021sequential}
K.~Liao, Y.~Zhao, J.~Gu, Y.~Zhang, and Y.~Zhong, ``Sequential convolutional
  recurrent neural networks for fast automatic modulation classification,''
  \emph{IEEE Access}, vol.~9, pp. 27\,182--27\,188, 2021.

\bibitem{ramjee2019fast}
S.~Ramjee, S.~Ju, D.~Yang, X.~Liu, A.~E. Gamal, and Y.~C. Eldar, ``Fast deep
  learning for automatic modulation classification,'' \emph{arXiv preprint
  arXiv:1901.05850}, 2019.

\bibitem{hou2022multi}
C.~Hou, G.~Liu, Q.~Tian, Z.~Zhou, L.~Hua, and Y.~Lin, ``Multi-signal modulation
  classification using sliding window detection and complex convolutional
  network in frequency domain,'' \emph{IEEE Internet of Things Journal}, 2022.

\bibitem{dong2022lightweight}
B.~Dong, Y.~Liu, G.~Gui, X.~Fu, H.~Dong, B.~Adebisi, H.~Gacanin, and H.~Sari,
  ``A lightweight decentralized learning-based automatic modulation
  classification method for resource-constrained edge devices,'' \emph{IEEE
  Internet of Things Journal}, 2022.

\bibitem{fu2022automatic}
X.~Fu, G.~Gui, Y.~Wang, H.~Gacanin, and F.~Adachi, ``Automatic modulation
  classification based on decentralized learning and ensemble learning,''
  \emph{IEEE Transactions on Vehicular Technology}, 2022.

\bibitem{dosovitskiy2020image}
A.~Dosovitskiy, L.~Beyer, A.~Kolesnikov, D.~Weissenborn, X.~Zhai,
  T.~Unterthiner, M.~Dehghani, M.~Minderer, G.~Heigold, S.~Gelly \emph{et~al.},
  ``An image is worth 16x16 words: Transformers for image recognition at
  scale,'' \emph{arXiv preprint arXiv:2010.11929}, 2020.

\bibitem{wang2021pyramid}
W.~Wang, E.~Xie, X.~Li, D.-P. Fan, K.~Song, D.~Liang, T.~Lu, P.~Luo, and
  L.~Shao, ``Pyramid vision transformer: A versatile backbone for dense
  prediction without convolutions,'' in \emph{Proceedings of the IEEE/CVF
  International Conference on Computer Vision}, 2021, pp. 568--578.

\bibitem{graham2021levit}
B.~Graham, A.~El-Nouby, H.~Touvron, P.~Stock, A.~Joulin, H.~J{\'e}gou, and
  M.~Douze, ``Levit: a vision transformer in convnet's clothing for faster
  inference,'' in \emph{Proceedings of the IEEE/CVF International Conference on
  Computer Vision}, 2021, pp. 12\,259--12\,269.

\bibitem{hamidi2021mcformer}
S.~Hamidi-Rad and S.~Jain, ``Mcformer: A transformer based deep neural network
  for automatic modulation classification,'' in \emph{2021 IEEE Global
  Communications Conference (GLOBECOM)}.\hskip 1em plus 0.5em minus 0.4em\relax
  IEEE, 2021, pp. 1--6.

\bibitem{goodfellow2014explaining}
I.~J. Goodfellow, J.~Shlens, and C.~Szegedy, ``Explaining and harnessing
  adversarial examples,'' \emph{arXiv preprint arXiv:1412.6572}, 2014.

\bibitem{sharif2016accessorize}
M.~Sharif, S.~Bhagavatula, L.~Bauer, and M.~K. Reiter, ``Accessorize to a
  crime: Real and stealthy attacks on state-of-the-art face recognition,'' in
  \emph{Proceedings of the 2016 acm sigsac conference on computer and
  communications security}, 2016, pp. 1528--1540.

\bibitem{xie2017adversarial}
C.~Xie, J.~Wang, Z.~Zhang, Y.~Zhou, L.~Xie, and A.~Yuille, ``Adversarial
  examples for semantic segmentation and object detection,'' in
  \emph{Proceedings of the IEEE international conference on computer vision},
  2017, pp. 1369--1378.

\bibitem{hendrik2017universal}
J.~Hendrik~Metzen, M.~Chaithanya~Kumar, T.~Brox, and V.~Fischer, ``Universal
  adversarial perturbations against semantic image segmentation,'' in
  \emph{Proceedings of the IEEE international conference on computer vision},
  2017, pp. 2755--2764.

\bibitem{jia2017adversarial}
R.~Jia and P.~Liang, ``Adversarial examples for evaluating reading
  comprehension systems,'' \emph{arXiv preprint arXiv:1707.07328}, 2017.

\bibitem{hu2017generating}
W.~Hu and Y.~Tan, ``Generating adversarial malware examples for black-box
  attacks based on gan,'' \emph{arXiv preprint arXiv:1702.05983}, 2017.

\bibitem{sadeghi2018adversarial}
M.~Sadeghi and E.~G. Larsson, ``Adversarial attacks on deep-learning based
  radio signal classification,'' \emph{IEEE Wireless Communications Letters},
  vol.~8, no.~1, pp. 213--216, 2018.

\bibitem{zhang2021countermeasures}
L.~Zhang, S.~Lambotharan, G.~Zheng, B.~AsSadhan, and F.~Roli, ``Countermeasures
  against adversarial examples in radio signal classification,'' \emph{IEEE
  Wireless Communications Letters}, vol.~10, no.~8, pp. 1830--1834, 2021.

\bibitem{lu2022Icassp}
L.~Zhang, S.~Lambotharan, and G.~Zheng, ``Adversarial learning in transformer
  based neural network in radio signal classification,'' in \emph{ICASSP 2022
  IEEE International Conference on Acoustics, Speech and Signal Processing
  (ICASSP)}.\hskip 1em plus 0.5em minus 0.4em\relax IEEE, 2022, pp. 1--5.

\bibitem{zhou2019access}
Z.~Zhou, Y.~Guo, Y.~He, X.~Zhao, and W.~M. Bazzi, ``Access control and resource
  allocation for m2m communications in industrial automation,'' \emph{IEEE
  Transactions on Industrial Informatics}, vol.~15, no.~5, pp. 3093--3103,
  2019.

\bibitem{liu2018deep}
M.~Liu, T.~Song, and G.~Gui, ``Deep cognitive perspective: Resource allocation
  for noma-based heterogeneous iot with imperfect sic,'' \emph{IEEE Internet of
  Things Journal}, vol.~6, no.~2, pp. 2885--2894, 2018.

\bibitem{zhou2015energy}
Z.~Zhou, M.~Dong, K.~Ota, G.~Wang, and L.~T. Yang, ``Energy-efficient resource
  allocation for d2d communications underlaying cloud-ran-based lte-a
  networks,'' \emph{IEEE Internet of Things Journal}, vol.~3, no.~3, pp.
  428--438, 2015.

\bibitem{tang2019future}
F.~Tang, Y.~Kawamoto, N.~Kato, and J.~Liu, ``Future intelligent and secure
  vehicular network toward 6g: Machine-learning approaches,'' \emph{Proceedings
  of the IEEE}, vol. 108, no.~2, pp. 292--307, 2019.

\bibitem{papernot2016distillation}
N.~Papernot, P.~McDaniel, X.~Wu, S.~Jha, and A.~Swami, ``Distillation as a
  defense to adversarial perturbations against deep neural networks,'' in
  \emph{2016 IEEE symposium on security and privacy (SP)}.\hskip 1em plus 0.5em
  minus 0.4em\relax IEEE, 2016, pp. 582--597.

\bibitem{jia2019comdefend}
X.~Jia, X.~Wei, X.~Cao, and H.~Foroosh, ``Comdefend: An efficient image
  compression model to defend adversarial examples,'' in \emph{Proceedings of
  the IEEE/CVF conference on computer vision and pattern recognition}, 2019,
  pp. 6084--6092.

\bibitem{kurakin2016adversarial}
A.~Kurakin, I.~Goodfellow, S.~Bengio \emph{et~al.}, ``Adversarial examples in
  the physical world,'' 2016.

\bibitem{ma2018characterizing}
X.~Ma, B.~Li, Y.~Wang, S.~M. Erfani, S.~Wijewickrema, G.~Schoenebeck, D.~Song,
  M.~E. Houle, and J.~Bailey, ``Characterizing adversarial subspaces using
  local intrinsic dimensionality,'' in \emph{ICLR 2018: Proceedings of the 6th
  International Conference on Learning Representations}.\hskip 1em plus 0.5em
  minus 0.4em\relax ICLR, 2018, pp. 1--15.

\bibitem{madry2017towards}
A.~Madry, A.~Makelov, L.~Schmidt, D.~Tsipras, and A.~Vladu, ``Towards deep
  learning models resistant to adversarial attacks,'' in \emph{International
  Conference on Learning Representations}, 2018.

\bibitem{athalye2018obfuscated}
A.~Athalye, N.~Carlini, and D.~Wagner, ``Obfuscated gradients give a false
  sense of security: Circumventing defenses to adversarial examples,'' in
  \emph{International conference on machine learning}.\hskip 1em plus 0.5em
  minus 0.4em\relax PMLR, 2018, pp. 274--283.

\bibitem{croce2020reliable}
F.~Croce and M.~Hein, ``Reliable evaluation of adversarial robustness with an
  ensemble of diverse parameter-free attacks,'' in \emph{International
  conference on machine learning}.\hskip 1em plus 0.5em minus 0.4em\relax PMLR,
  2020, pp. 2206--2216.

\bibitem{goldblum2020adversarially}
M.~Goldblum, L.~Fowl, S.~Feizi, and T.~Goldstein, ``Adversarially robust
  distillation,'' in \emph{Proceedings of the AAAI Conference on Artificial
  Intelligence}, vol.~34, no.~04, 2020, pp. 3996--4003.

\bibitem{zhu2021reliable}
J.~Zhu, J.~Yao, B.~Han, J.~Zhang, T.~Liu, G.~Niu, J.~Zhou, J.~Xu, and H.~Yang,
  ``Reliable adversarial distillation with unreliable teachers,'' \emph{arXiv
  preprint arXiv:2106.04928}, 2021.

\bibitem{chen2020robust}
T.~Chen, Z.~Zhang, S.~Liu, S.~Chang, and Z.~Wang, ``Robust overfitting may be
  mitigated by properly learned smoothening,'' in \emph{International
  Conference on Learning Representations}, 2020.

\bibitem{zi2021revisiting}
B.~Zi, S.~Zhao, X.~Ma, and Y.-G. Jiang, ``Revisiting adversarial robustness
  distillation: Robust soft labels make student better,'' in \emph{Proceedings
  of the IEEE/CVF International Conference on Computer Vision}, 2021, pp.
  16\,443--16\,452.

\bibitem{ba2016layer}
J.~L. Ba, J.~R. Kiros, and G.~E. Hinton, ``Layer normalization,'' \emph{arXiv
  preprint arXiv:1607.06450}, 2016.

\bibitem{he2016deep}
K.~He, X.~Zhang, S.~Ren, and J.~Sun, ``Deep residual learning for image
  recognition,'' in \emph{Proceedings of the IEEE conference on computer vision
  and pattern recognition}, 2016, pp. 770--778.

\bibitem{vaswani2017attention}
A.~Vaswani, N.~Shazeer, N.~Parmar, J.~Uszkoreit, L.~Jones, A.~N. Gomez,
  {\L}.~Kaiser, and I.~Polosukhin, ``Attention is all you need,''
  \emph{Advances in neural information processing systems}, vol.~30, 2017.

\bibitem{Biggio2018}
B.~Biggio and F.~Roli, ``{Wild patterns: Ten years after the rise of
  adversarial machine learning},'' \emph{Pattern Recognition}, vol.~84, pp.
  317--331, 2018.

\bibitem{OShea2016}
T.~J. O'Shea and N.~West, ``{Radio Machine Learning Dataset Generation with GNU
  Radio},'' \emph{Proceedings of the GNU Radio Conference}, vol.~1, no.~1,
  2016.

\bibitem{luan2022automatic}
S.~Luan, Y.~Gao, T.~Liu, J.~Li, and Z.~Zhang, ``Automatic modulation
  classification: Cauchy-score-function-based cyclic correlation spectrum and
  fc-mlp under mixed noise and fading channels,'' \emph{Digital Signal
  Processing}, vol. 126, p. 103476, 2022.

\bibitem{papernot2016transferability}
N.~Papernot, P.~McDaniel, and I.~Goodfellow, ``Transferability in machine
  learning: from phenomena to black-box attacks using adversarial samples,''
  \emph{arXiv preprint arXiv:1605.07277}, 2016.

\end{thebibliography}
}

\vspace*{-2\baselineskip}

\begin{IEEEbiography}[{\includegraphics[width=1in,height=1.25in,clip,keepaspectratio]{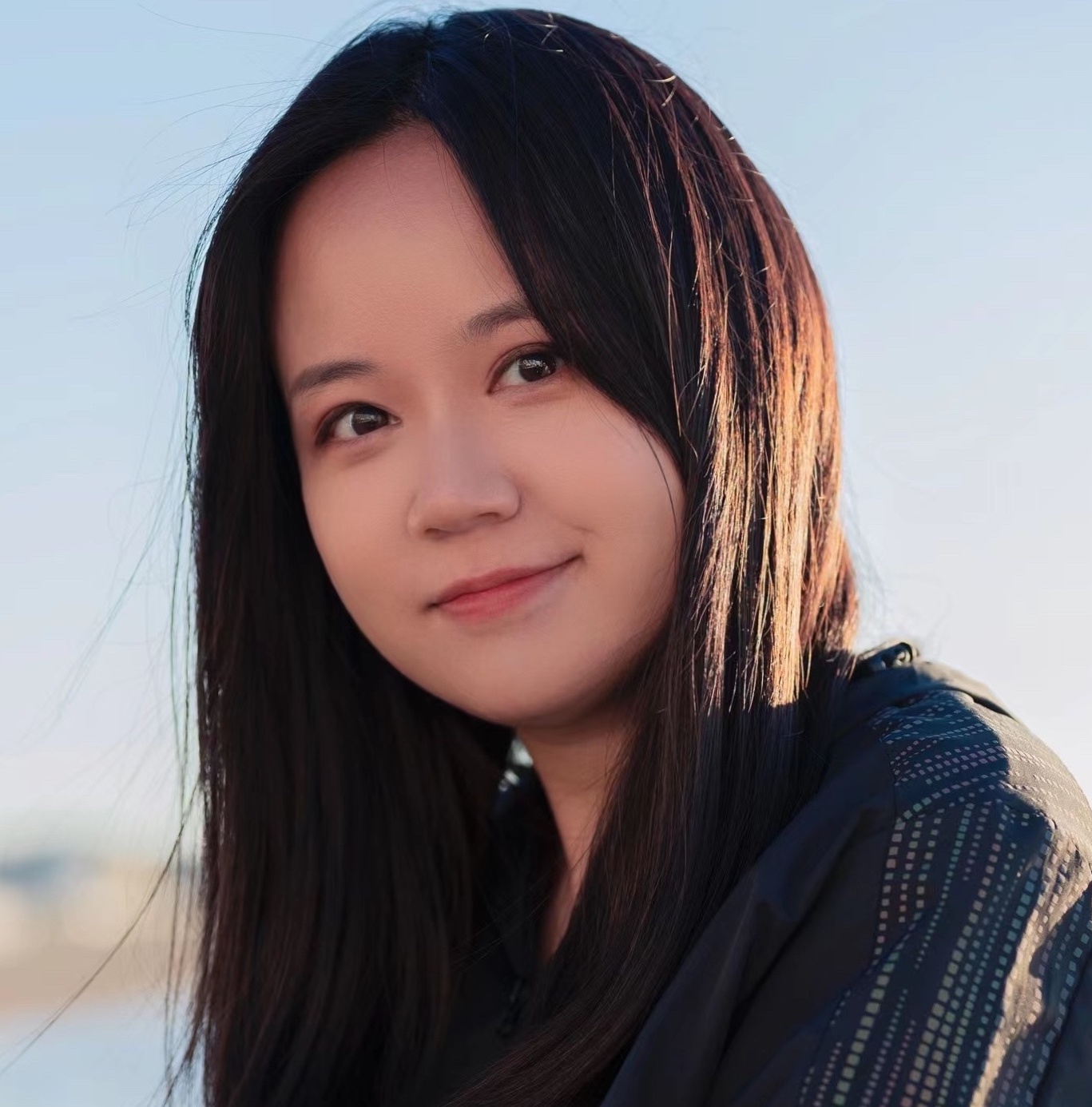}}]{Lu Zhang} is a Ph.D. student in the Signal Processing and Networks Research Group at Loughborough University, UK. She received B.Eng in Electronic and Information Engineering from Xidian University, China and M.Sc (with distinction) in Mobile Communications from Loughborough University, UK, in 2016 and 2018, respectively. She received Clarke-Griffiths Best Student Prize from Loughborough University in 2018. Her research interests include deep learning and adversarial learning.
\end{IEEEbiography}
\vspace*{-3\baselineskip}
\begin{IEEEbiography}[{\includegraphics[width=1in,height=1.25in,clip,keepaspectratio]{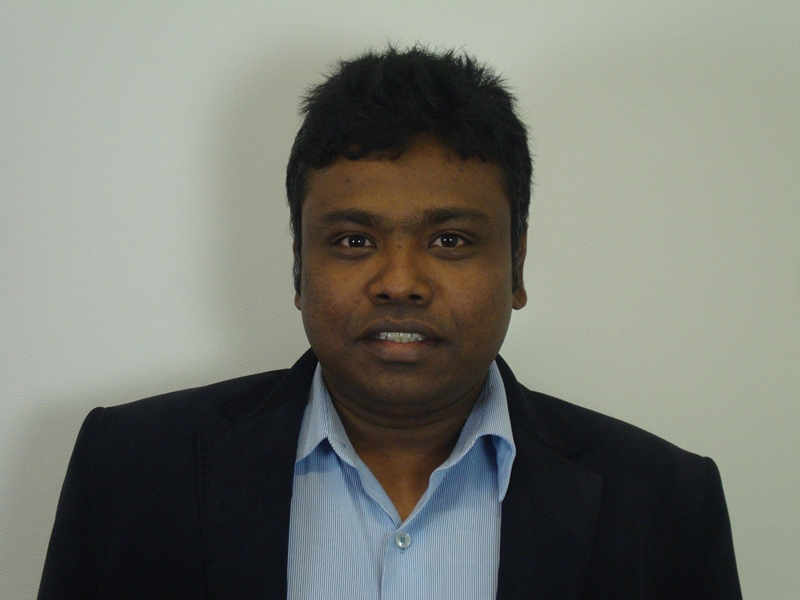}}]{Sangarapillai Lambotharan} received the Ph.D. degree in signal processing from Imperial College London, U.K., in 1997. He was a Visiting Scientist with the Engineering and Theory Centre, Cornell University, USA, in 1996. Until 1999, he was a Post-Doctoral Research Associate with Imperial College London. From 1999 to 2002, he was with the Motorola Applied Research Group, U.K., where he investigated various projects, including physical link layer modeling and performance characterization of GPRS, EGPRS, and UTRAN. He was with King’s College London and Cardiff University as a Lecturer and a Senior Lecturer, respectively, from 2002 to 2007. He is currently a Professor of Digital Communications and the Head of the Signal Processing and Networks Research Group, Wolfson School of Mechanical, Electrical and Manufacturing Engineering, Loughborough University, U.K. His current research interests include 5G networks, MIMO, blockchain, machine learning, and network security. He has authored more than 250 journal articles and conference papers in these areas. He is a Fellow of IET and Senior Member of IEEE. He serves as an Associate Editor for the IEEE Transactions on Signal Processing and IEEE Transactions on Communications.
\end{IEEEbiography}
\vspace*{-3\baselineskip}

\begin{IEEEbiography}[{\includegraphics[width=1in,height=1.25in,clip,keepaspectratio]{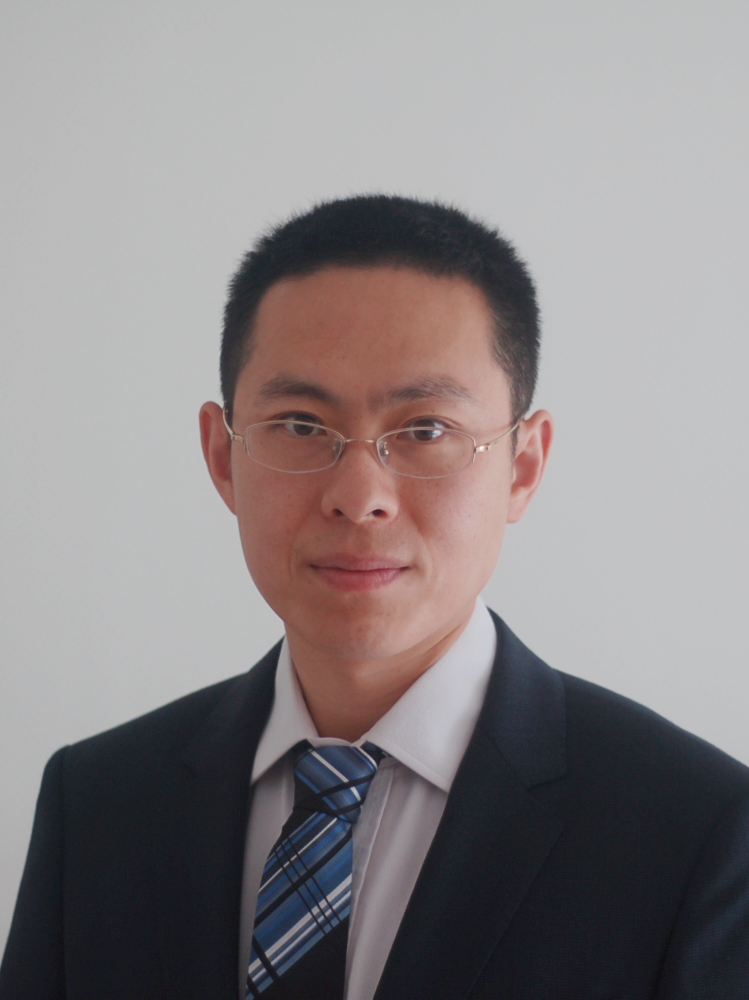}}]{Gan Zheng} (S'05-M'09-SM'12-F'21)  received the BEng and the MEng from Tianjin University, Tianjin, China, in 2002 and 2004, respectively, both in Electronic and Information Engineering, and the PhD degree in Electrical and Electronic Engineering from The University of Hong Kong in 2008. He is currently Professor in Connected Systems in the School of Engineering, University of Warwick, UK. His research interests include machine learning for wireless communications, UAV communications, mobile edge caching, full-duplex radio, and wireless power transfer. He is the first recipient for the 2013 IEEE Signal Processing Letters Best Paper Award, and he also received 2015 GLOBECOM Best Paper Award, and 2018 IEEE Technical Committee on Green Communications \& Computing Best Paper Award. He was listed as a Highly Cited Researcher by Thomson Reuters/Clarivate Analytics in 2019. He currently serves as an Associate Editor for IEEE Wireless Communications Letters.
\end{IEEEbiography}
\vspace*{-3\baselineskip}
\begin{IEEEbiography}[{\includegraphics[width=1in,height=1.25in,clip,keepaspectratio]{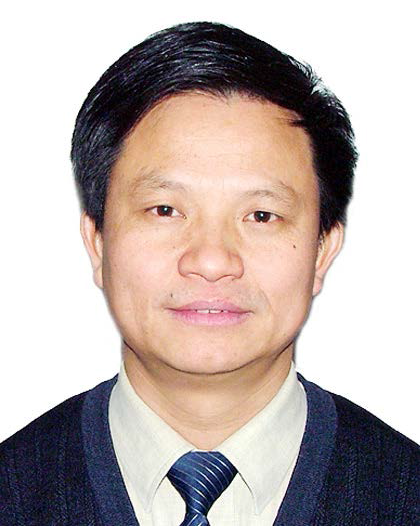}}]{Guisheng Liao} was born in Guangxi, China, in 1963. He received the B.S. degree in mathematics from Guangxi University, Guangxi, China, in 1985, the M.S. degree in computer software from Xidian University, Xi’an, China, in 1990, and the Ph.D. degree in signal and information processing from Xidian University, Xi’an, China, in 1992. From 1999 to 2000, he was a Senior Visiting Scholar with The Chinese University of Hong Kong, Hong Kong. Since 2006, he has been serving as the panelist for the medium- and long-term development plans in high-resolution and remote sensing systems. Since 2007, he has been the Lead of the Chang Jiang Scholars Innovative Team, Xidian University, and devoted to advanced techniques in signal and information processing. Since 2009, he has been the Evaluation Expert for the International Cooperation Project of the Ministry of Science and Technology in China. He was a Chang Jiang Scholars Distinguished Professor with the National Laboratory of Radar Signal Processing and serves as the Dean of the School of Electronic Engineering, Xidian University. He is the author or a coauthor of several books and more than 200 publications. His research interests include array signal processing, space–time adaptive processing, radar waveform design, and airborne/space surveillance and warning radar systems.
\end{IEEEbiography}
\vspace*{-3\baselineskip}
\begin{IEEEbiography}[{\includegraphics[width=1in,height=1.25in,clip,keepaspectratio]{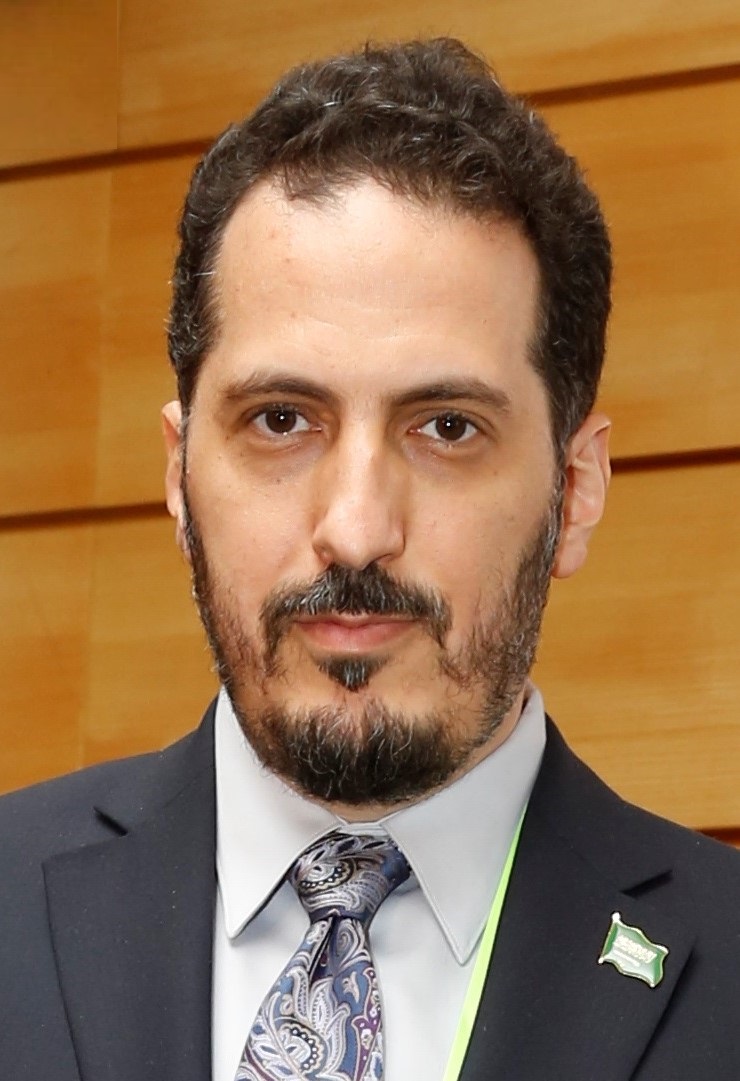}}]{Basil AsSadhan} received the M.Sc. degree in electrical and computer engineering from the University of Wisconsin and the Ph.D. degree in electrical and computer engineering from Carnegie Mellon University. He is an Associate Professor at the Electrical Engineering Department at King Saud University and an IEEE senior member. His research interests are in the areas of cybersecurity, network security, network traffic analysis, anomaly detection, and machine learning.
\end{IEEEbiography}
\vspace*{-3\baselineskip}
\begin{IEEEbiography}[{\includegraphics[width=1in,height=1.25in,clip,keepaspectratio]{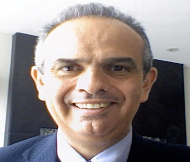}}]{Fabio Roli} is a Full Professor of Computer Engineering at the University of Genova, Italy, and Founding Director of the Pattern Recognition and Applications laboratory at the University of Cagliari. He has been doing research on the design of pattern recognition and machine learning systems for thirty years. He has provided seminal contributions to the fields of multiple classifier systems and adversarial machine learning, and he has played a leading role in the establishment and advancement of these research themes. He has been appointed Fellow of the IEEE and Fellow of the International Association for Pattern Recognition. He is a recipient of the Pierre Devijver Award for his contributions to statistical pattern recognition.
\end{IEEEbiography}

\end{document}